%% file: iclr2027_conference.tex
\documentclass[11pt,a4paper,logo,onecolumn]{trajlong_report}
\input{config/preamble}
\input{config/metadata}
\begin{abstract}
Training environments remain a major bottleneck for agents in data-driven scientific discovery. Existing collections of executable and verifiable training environments are small, and their construction remains difficult to scale. To address this limitation, we introduce \textbf{\poseidon{}}, an automated pipeline for generating diverse, verifiable training environments from scientific code repositories at scale. From the same repository budget, \poseidon{} extracts nearly \textbf{10$\times$} as many training environments as a prior approach by decomposing scientific workflows into four complementary task types: Reproduction, Subflow, Completion, and Correction, where each type targets a distinct capability for scientific research. A recursive improvement loop further refines failed candidates and recovers otherwise discarded environments. Using \poseidon{}, we construct \textbf{\pearl{}}, a collection of \textbf{97,659} executable and verifiable environments from \textbf{7,110} public repositories spanning \textbf{10} scientific disciplines. To the best of our knowledge, \pearl{} is the largest and most broadly multidisciplinary collection of scientific-agent training environments to date. Qwen3.5 models (9B, 27B, and 35B-A3B) trained on \pearl{} achieve consistent gains across six challenging scientific-agent benchmarks. Overall, the scale and diversity of \pearl{} offer opportunities for training scientific discovery agents. 
\end{abstract}

\begin{document}
\maketitle

\begin{figure}[h]
    \centering
    \includegraphics[width=\linewidth]{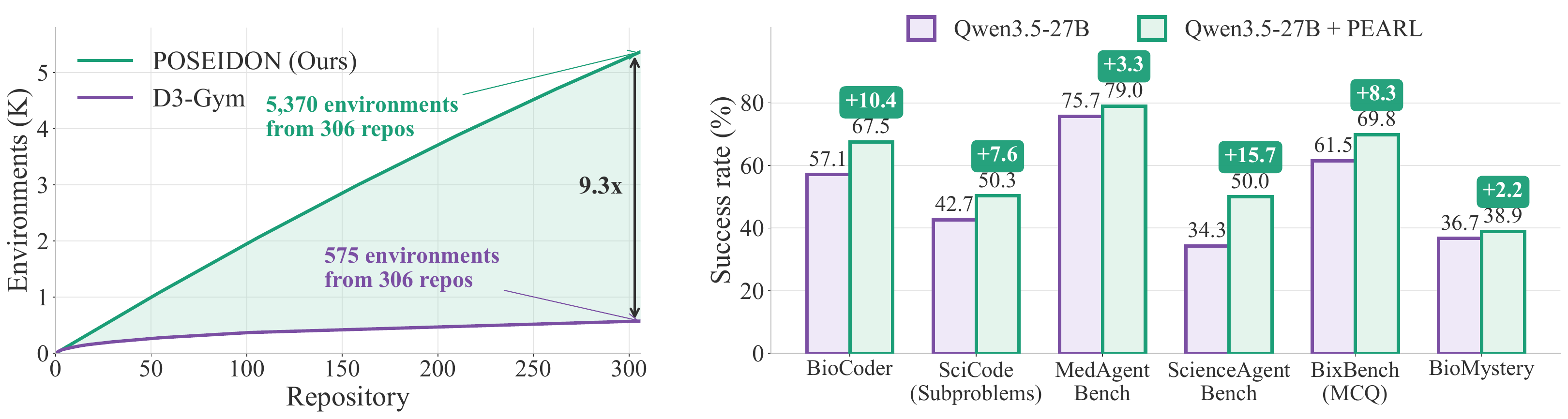}
\caption{
\textbf{Scaling scientific-agent training environments and improving downstream performance.}
\textbf{Left:} Cumulative number of environments synthesized as the number of source repositories increases. Under a matched repository budget of 306 repositories, \poseidon{} produces 5,370 environments, compared with 575 produced by D3-Gym---a 9.3$\times$ higher yield. 
\textbf{Right:} Success rates of Qwen3.5-27B before and after training on \pearl{} across six scientific-discovery benchmarks. 
}
    \vspace{-0.5cm}
    \label{fig:makeb}
    \end{figure}

\input{sections/01_introduction_v1}
\input{sections/02_related_work}
\input{sections/03_method}

\input{sections/04_experiment}
\input{sections/05_conclusion}

\clearpage
\bibliography{iclr2027_conference}
\bibliographystyle{iclr2027_conference}

\clearpage
\appendix
\input{sections/A1_appendix_head}
\input{sections/A1_5_verification}
\input{sections/A1_6_human_verification}
\input{sections/A2_benchmark_details}
\input{sections/A3_data_examples}
\input{sections/A4_prompts}
\clearpage
\input{sections/A5_dataset_comparison}
\end{document}

%% file: config/preamble.tex
\usepackage[authoryear,sort&compress,round]{natbib}
\usepackage{microtype}
\definecolor{myCite}{HTML}{1C4587}
\hypersetup{citecolor=myCite,linkcolor=myCite,urlcolor=myCite}
\usepackage{placeins}
\usepackage{multirow}
\usepackage{threeparttable}
\usepackage{makecell}
\usepackage{tabularx}
\usepackage[table]{xcolor}
\usepackage{pifont}
\usepackage[most]{tcolorbox}
\usepackage{listings}
\usepackage{wrapfig}
\usepackage{adjustbox}
\usepackage{mathtools}
\usepackage{subcaption}

\IfFileExists{OT1XCharter-TLF.fd}{%
  \SetSymbolFont{operators}{normal}{OT1}{XCharter-TLF}{m}{n}
  \SetSymbolFont{operators}{bold}{OT1}{XCharter-TLF}{b}{n}
  \SetMathAlphabet{\mathit}{normal}{OT1}{XCharter-TLF}{m}{it}
  \SetMathAlphabet{\mathit}{bold}{OT1}{XCharter-TLF}{b}{it}
  \SetMathAlphabet{\mathbf}{normal}{OT1}{XCharter-TLF}{b}{n}
  \SetMathAlphabet{\mathbf}{bold}{OT1}{XCharter-TLF}{b}{n}
}{}
\input{math_commands.tex}

\definecolor{basecolor}{HTML}{D4DFF7}
\definecolor{frontiercolor}{HTML}{F1E6FA}
\definecolor{baseline9color}{HTML}{E9F9E6}
\definecolor{baseline27color}{HTML}{FCECEC}
\definecolor{ourscolor}{HTML}{FFF2CC}

\newcommand{\yes}{\textcolor{green!60!black}{\ding{51}}}
\newcommand{\no}{\textcolor{red!70!black}{\ding{55}}}
\newcolumntype{L}[1]{>{\raggedright\arraybackslash}p{#1}}
\newcolumntype{C}[1]{>{\centering\arraybackslash}p{#1}}
\providecommand{\method}{Pearl}
\definecolor{oasisgreen}{RGB}{30, 130, 76}
\newcommand{\poseidon}{\textsc{Poseidon}}
\newcommand{\pearl}{\textsc{Pearl}}

\newtcbtheorem[number within=section]{exmp}{Prompts}{breakable,colback=white!5!white,colframe=black!95!,fonttitle=\bfseries,left=.02in,right=.02in,bottom=.02in,top=.02in}{exmp}
\newtcbtheorem[number within=section]{detail}{Problem Description}{breakable,colback=white!5!white,colframe={rgb,255:red,78; green,136; blue,114},fonttitle=\bfseries,left=.02in,right=.02in,bottom=.02in,top=.02in}{detail}
\definecolor{mybarcolor}{HTML}{c7bde5}
\newmdenv[topline=false,bottomline=false,rightline=false,skipabove=10pt,skipbelow=0pt,linewidth=3pt,linecolor=mybarcolor,innerleftmargin=10pt,innerrightmargin=0pt,innertopmargin=0pt,innerbottommargin=0pt,innermargin=0pt]{leftbar}

\newcommand{\logo}[1]{\raisebox{-0.22\height}{\includegraphics[height=1.1em]{logos/#1}}\hspace{2pt}}
\definecolor{gaincolor}{HTML}{1B7F3B}
\definecolor{dropcolor}{HTML}{C0392B}
\newcommand{\gain}[1]{\textcolor{gaincolor}{\scriptsize\textbf{$\uparrow$#1}}}
\newcommand{\drop}[1]{\textcolor{dropcolor}{\scriptsize\textbf{$\downarrow$#1}}}
\definecolor{bestcell}{RGB}{214, 223, 245}

%% file: math_commands.tex
\usepackage{amsmath,amsfonts,bm}

\def\eqref#1{equation~\ref{#1}}

\def\1{\bm{1}}

\DeclareMathAlphabet{\mathsfit}{\encodingdefault}{\sfdefault}{m}{sl}
\SetMathAlphabet{\mathsfit}{bold}{\encodingdefault}{\sfdefault}{bx}{n}



%% file: config/metadata.tex
\title{POSEIDON: Scaling Training Environments for Scientific Agents from Code Repositories}

\renewcommand{\reportlogos}{%
  \includegraphics[height=17.05pt]{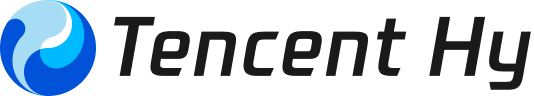}%
  \hspace{12pt}%
  \includegraphics[height=17.05pt]{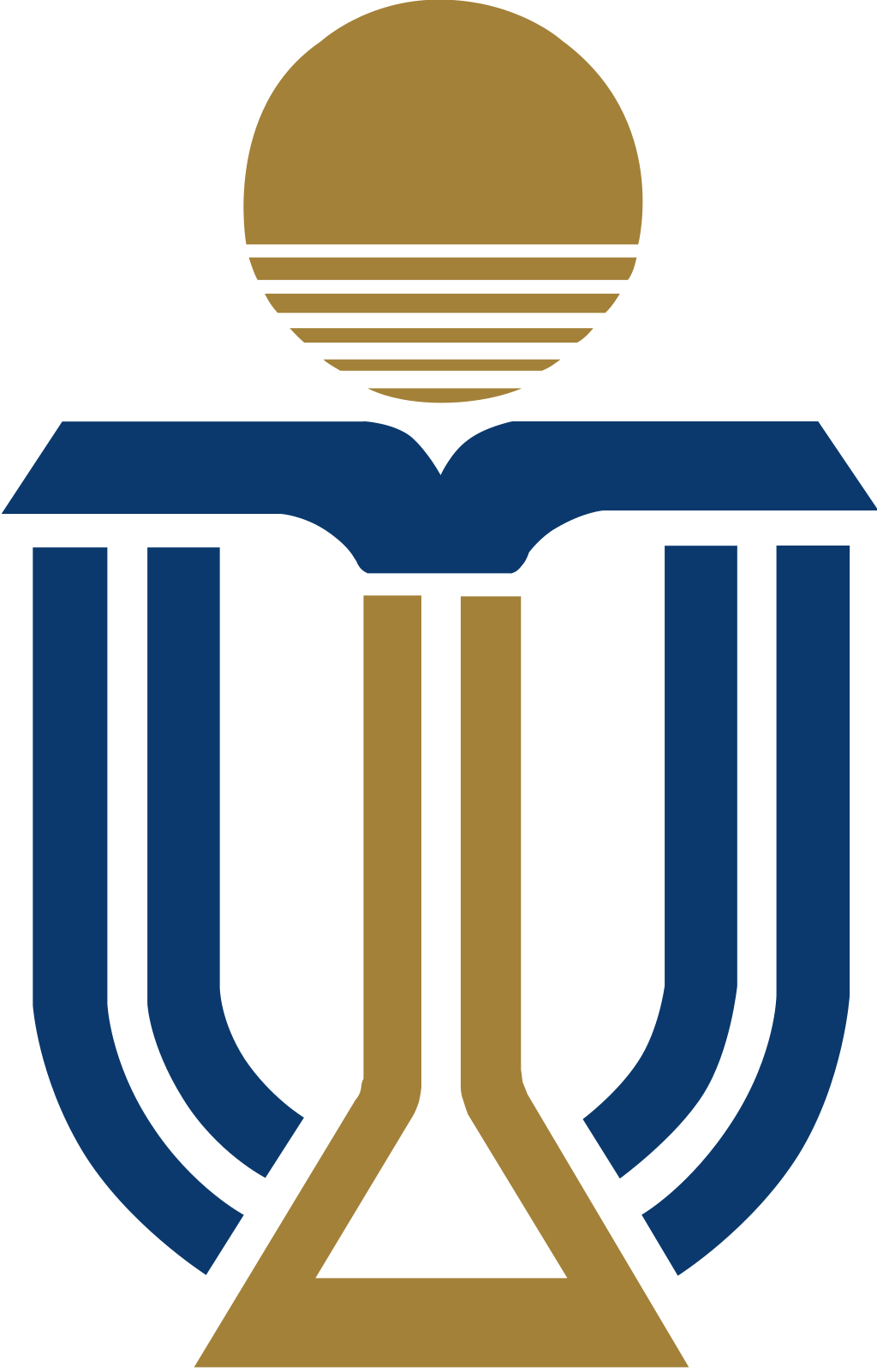}%
}

\DeclareRobustCommand{\paperauthorblock}{%
  \fontsize{10.5}{14}\selectfont
  Qifan Zhang\textsuperscript{1,2*},
  Yuhan Li\textsuperscript{2\textdagger},
  Nuo Chen\textsuperscript{2\textdaggerdbl},
  Bowen Liu\textsuperscript{2*},
  Xiao Hu\textsuperscript{2},
  Miao Peng\textsuperscript{1,2}\\
  Chenyi Zi\textsuperscript{1},
  Lixin Liu\textsuperscript{2},
  Lydell Huang\textsuperscript{2},
  Jing Tang\textsuperscript{1},
  Jia Li\textsuperscript{1\textdagger}\\[4pt]
  \normalfont\fontsize{9.5}{12}\selectfont
  \textsuperscript{1}The Hong Kong University of Science and Technology (Guangzhou)\\
  \textsuperscript{2}Hunyuan Team, Tencent%
}
\author{\paperauthorblock}
\AddToHook{cmd/maketitle/after}{%
  \begingroup
  \footnotetext[1]{Equal contribution.}%
  \footnotetext[3]{Project Lead.}%
  \footnotetext[2]{Corresponding authors.}%
  \endgroup
}
\hypersetup{pdfauthor={Qifan Zhang, Yuhan Li, Nuo Chen, Bowen Liu, Xiao Hu, Miao Peng, Chenyi Zi, Lixin Liu, Lydell Huang, Jing Tang, Jia Li}}

%% file: sections/01_introduction_v1.tex
\section{Introduction}

\begin{wrapfigure}{r}{0.5\textwidth}
    \centering
    \vspace{-12pt}
    \includegraphics[width=\linewidth]{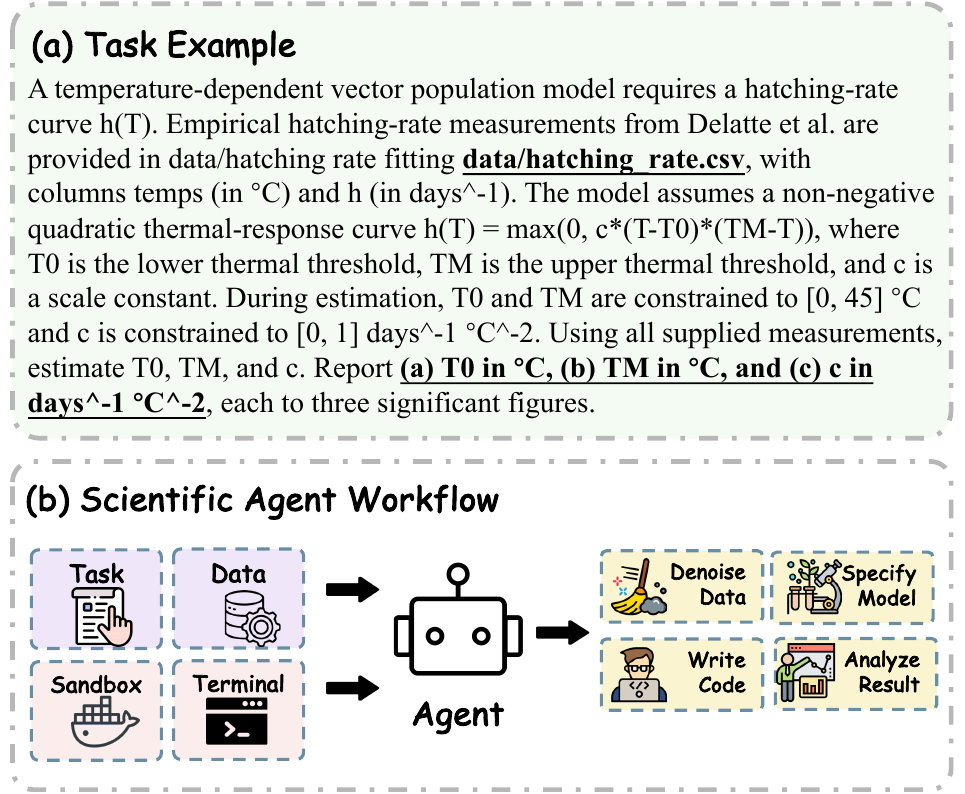}
    \caption{An example task from \pearl{} and the corresponding scientific-agent workflow.}
    \label{fig:data_example}
    \vspace{-10pt}
\end{wrapfigure}

The rapid growth of scientific data and the increasing use of large language model agents in research have sparked a surge of interest in data-driven scientific discovery~\citep{xin2025towards,gridach2025agentic,anthropic2026biomysterybench,majumder2024discoverybench, liu2026lifescibench}. Unlike knowledge-based question answering or isolated code generation, scientific research requires agents to carry out \textbf{scientific workflows}: coordinated sequences of data processing and computational steps that address a research objective. Agents must clean data, select appropriate models, write and debug programs, and iteratively revise their analyses based on intermediate results before reaching evidence-supported conclusions~\citep{schmidgall2025agentlaboratoryusingllm,zhou2026autonomousagentsscientificdiscovery} (as depicted in Figure~\ref{fig:data_example}). The capability to carry out these scientific workflows requires integrating \textbf{scientific reasoning with coding and tool use}, and remains difficult for even current frontier models to master. For example, on OpenAI's GeneBench-Pro~\citep{li2026genebench}, GPT-5.6 Sol~\citep{openai2026gpt56} achieves a pass rate of only 28.7\% at the highest reasoning level (or 31.5\% with Pro mode enabled), highlighting the immense difficulty of reliably automating complex scientific analyses. Yet, \textbf{training environments designed to develop this capability remain severely limited.}

D3-Gym~\citep{moussa2026d3} pioneered the automated synthesis of such training environments from scientific code repositories. These repositories are valuable because they encode \textbf{the scientific workflows described above} in related sets of code files that connect research objectives, input data, and computational procedures. Executing these workflows provides reference outputs for verification. However, \textbf{scaling this approach remains challenging}. D3-Gym converts each extracted workflow into an end-to-end task, leaving more varied supervision within the workflow underused. Its collection contains only 565 environments from 239 repositories (approximately 2.4 environments per repository). This limited yield raises a central research question:

\begin{leftbar}
\textit{How can we efficiently exploit executable scientific workflows within code repositories to generate high-quality, diverse training environments at scale?}
\end{leftbar}

To answer this, we introduce \textbf{POSEIDON}, an automated pipeline for mining scientific code repositories through three stages: \textit{source discovery}, \textit{task synthesis}, and \textit{verification}. Source discovery identifies repositories containing scientific code, data, and research context through domain-keyword expansion. Task synthesis extracts scientific workflows and derives four complementary task types, each corresponding to a distinct capability for scientific agents: \texttt{Reproduction}, reproducing the results of a complete scientific workflow from its research objective; \texttt{Subflow}, carrying out a standalone analysis to answer a focused scientific question motivated by a broader research context; \texttt{Completion}, filling in a missing function in the provided workflow code; and \texttt{Correction}, identifying and fixing an error in provided code that produces incorrect results. Our key insight is that \textbf{a scientific workflow can provide not only an end-to-end research task, but also richer supervision through structured modifications to its constituent steps}. Under a matched budget of 306 repositories, \textbf{POSEIDON produces 5,370 environments, 9.3$\times$ as many as the 575 produced by D3-Gym}, as shown in the left panel of Figure~\ref{fig:makeb}.

In the verification stage, we introduce a \textbf{recursive improvement} mechanism driven by an iterative generator--validator loop. If a generated environment fails the initial reliability and scientific-validity checks, the validator returns diagnostic feedback and revision suggestions, after which the generator revises and resubmits the environment. Among the environments rejected in the first verification round, \textbf{49.4\%} are successfully recovered and pass verification through this refinement process. This closed-loop refinement recovers flawed but fixable candidates that would otherwise be discarded, thereby improving environment quality while increasing data yield.

Using POSEIDON, we construct \textbf{PEARL}, a large-scale collection of verified scientific-agent training environments sourced from \textbf{7,110 repositories across 10 scientific disciplines}. PEARL contains \textbf{97,659 environments}, each equipped with a clear task description, input data, an executable sandbox, a reference solution, and an automatic evaluation script. \textbf{To the best of our knowledge, PEARL is the largest and most broadly multidisciplinary collection of training environments for scientific discovery agents to date}. We evaluate PEARL through supervised fine-tuning on Qwen3.5-9B, 27B, and 35B-A3B~\citep{qwen3.5}, achieving notable improvements across six highly challenging scientific-agent benchmarks. Furthermore, controlled ablations show that both broader repository coverage and complementary workflow-derived task types contribute to downstream performance. Overall, the scale and diversity of PEARL open up new opportunities for training scientific discovery agents.

%% file: sections/02_related_work.tex
\section{Related Works}
\label{others}

\paragraph{Scalable Task Synthesis from Code Repositories.}
Executable code repositories have been extensively leveraged as rich sources of training supervision. Prior work derives completion tasks by masking code at multiple granularities or isolating target functions and their dependencies~\citep{zhang2023repocoder,yang2024execrepobench,xie2025repost}, and converts naturally occurring code fragments into executable specifications and test cases~\citep{xie2024codebenchgen}. For software repair, scalable pipelines synthesize verifiable tasks by injecting test-breaking defects via language-model rewrites and procedural mutations, or by simulating repository-level debugging workflows~\citep{yang2025swesmith, pham2025swe}. Other methods reconstruct executable issue-resolution tasks from repository histories through test generation, back-translation, patch reversal, and code-state reconstruction~\citep{jain2025r2egym,qi2026change2task}. More recently, capability-targeted tasks, including feature development, bug fixing, function localization, and dependency search, have been derived from shared repository contexts~\citep{fu2026corecodebench,xie2026hybridgym}. Together, these studies demonstrate that transforming and restructuring existing code artifacts can substantially increase both the quantity and diversity of supervision extracted from a single repository.\textbf{ Inspired by this line of work}, we decompose and transform executable scientific workflows into four complementary task types, thereby scaling the generation of training environments for data-driven scientific discovery.

\paragraph{Automated Training Environment Synthesis for Scientific Discovery.}
With the rapid growth of scientific data and the increasing capabilities of LLM agents, data-driven scientific discovery has attracted growing attention, spanning scientific coding, computational reproducibility, and autonomous data analysis~\citep{li2026genebench,anthropic2026biomysterybench,mitchener2025bixbench,chen2025scienceagentbench,liu2026lifescibench}. However, large-scale executable environments for training agents to conduct such workflows remain scarce. D3-Gym~\citep{moussa2026d3} constructs verifiable training environments from scientific repositories by treating extracted scientific workflows as end-to-end tasks. This approach produces relatively few environments per repository, making it challenging to scale. \textbf{In contrast, we derive multiple complementary tasks from each executable scientific workflow}, providing richer supervision beyond end-to-end execution and enabling the generation of diverse, verifiable training environments at scale.

%% file: sections/03_method.tex
\section{Method}

In this work, we aim to enhance the problem-solving capabilities of scientific agents by increasing the scale and diversity of training environments. To achieve this, we introduce POSEIDON, an automated synthesis pipeline, and use it to construct PEARL, a large-scale, verified collection of such environments. We begin by formally defining the scientific agent task, and then detail the three stages of the POSEIDON pipeline: repository discovery, task synthesis, and verification.

\begin{figure}[t]
    \centering
    \includegraphics[width=\linewidth]{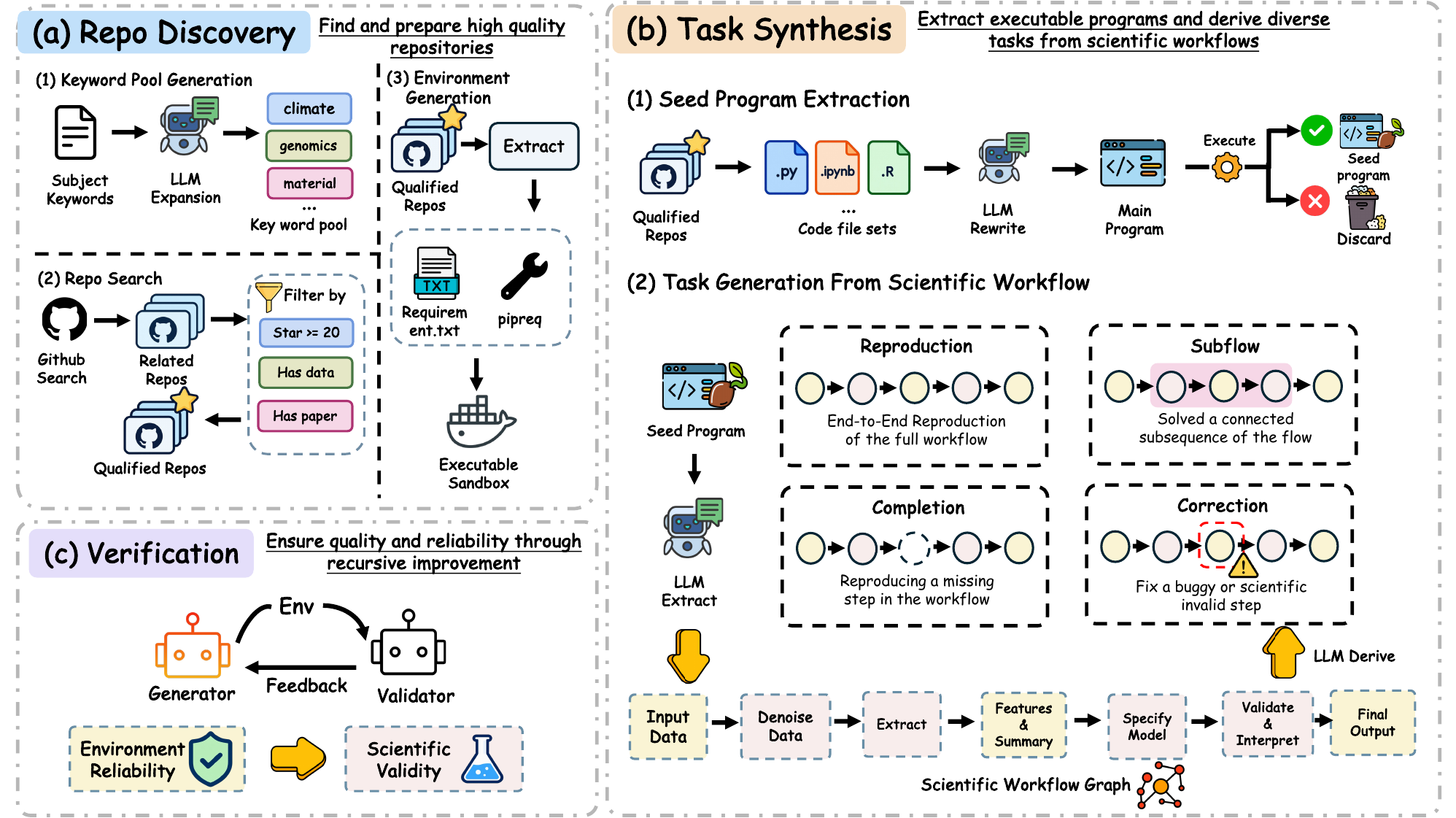}
    \caption{Overview of \poseidon{}}
    \vspace{-0.5cm}
    \label{fig:method}
    \end{figure}

\subsection{Problem Formulation}

\paragraph{SciAgent.}
We study scientific agents (SciAgents). Given a task description, an input data, an executable sandbox, and a code interpreter tool, the agent must denoise the data, choose suitable models, write code, analyse execution results, and produce a final answer. We formulate this process as a sequential decision problem. An agent policy $\pi_{\theta}$ performs interleaved reasoning and tool use. At step $t$, the agent generates a thought $r_t$ and an action $a_t$ based on the history; the environment executes $a_t$ and returns an observation $o_t$. This yields a trajectory $\tau = (r_1, a_1, o_1, \ldots, r_T, a_T, o_T)$. 
Figure~\ref{fig:data_example} illustrates a representative task together with the corresponding SciAgent workflow.

\paragraph{Environment Synthesis Objectives.}
To support reliable and effective agent training, SciAgent training environment should satisfy four properties:
\textbf{(1) Data-Dependent:} Each environment must be grounded in at least one input dataset that is necessary for solving the scientific problem.
\textbf{(2) Executable:} Each environment must provide a task-specific sandbox containing the dependencies required by the agent, ensuring at least one executable solution path.
\textbf{(3) Verifiable:} Each environment must include an evaluation script that determines whether the agent's output is correct.
\textbf{(4) Diverse:} The environments should span a broad range of source repositories and scientific disciplines, and cover diverse task types.

Formally, we represent a SciAgent training environment as
\[
\mathbb{E}
=
\left\{
\left(T_i, D_i, S_i, V_i\right)
\right\}_{i=1}^{N},
\]
where $T_i$ is the task description, $D_i$ is the input dataset, $S_i$ is the executable sandbox, and $V_i$ is the evaluation script.

To satisfy these objectives at scale, we select \textbf{GitHub}\footnote{\url{https://github.com}} as our primary data source. GitHub hosts millions of open-source repositories that naturally \textbf{bundle scientific code, environment configurations, and associated data}. This rich ecosystem provides an ideal foundation for extracting diverse, data-dependent, and executable workflows. Leveraging this resource, we design an automated pipeline to systematically transform these repositories into agent training environment.

\subsection{Environment Generation Pipeline}
The \poseidon{} pipeline turns public scientific repositories into training environment. It has three stages: \textbf{repository discovery, task synthesis, and verification.}

\subsubsection{Repository Discovery}
The scale and diversity of synthesized scientific tasks depend directly on the breadth and quality of their source repositories. We therefore first construct a large and diverse repository pool and build an executable sandbox for each repository, providing a reliable foundation for subsequent task synthesis (\textbf{Panel (a) of Figure~\ref{fig:method}}).

\paragraph{Keyword Pool Generation.}
We begin with a small set of domain descriptions, such as \texttt{structural biology and molecular dynamics} or \texttt{cheminformatics and drug discovery}. For each domain, we prompt an LLM to propose concrete software, libraries, file formats, and database names typically mentioned in the README files of related repositories, such as \texttt{GROMACS}, \texttt{MDAnalysis}, and \texttt{PLUMED} for molecular dynamics. Each candidate term is then executed as a GitHub search query. We discard terms with fewer than 30 hits, as they indicate limited public code availability, and terms with more than 5{,}000 hits, as they are often too generic (e.g., \texttt{docker} or \texttt{hdf5}). The retained terms constitute the \textbf{keyword pool}.

\begin{wrapfigure}{r}{0.46\textwidth}
    \centering
    \vspace{-12pt}
    \includegraphics[width=\linewidth]{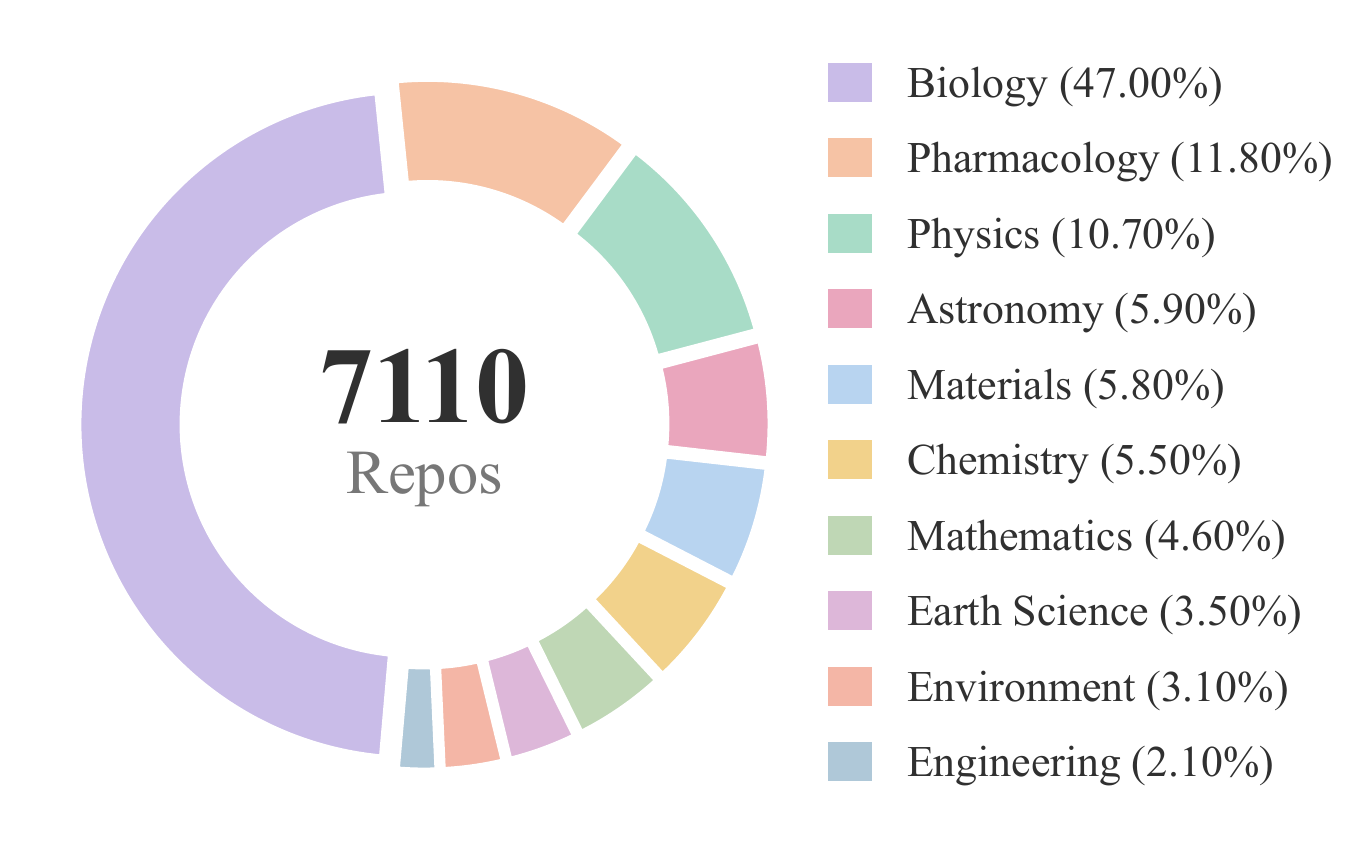}
    \caption{Distribution of the collected scientific repositories across disciplines.}
    \label{fig:discipline-pie}
    \vspace{-10pt}
\end{wrapfigure}

\paragraph{Repository Pool Collection.}
We query GitHub using the keyword pool and retain a repository only if it meets four criteria: \textbf{it has at least 20 stars, contains scientific data, is backed by a published paper or preprint, and belongs to a core scientific discipline} (e.g., mathematics, physics, chemistry, biology, or pharmacy). The retained repositories form the \textbf{repository pool}.

\paragraph{Executable Sandbox Construction.}
For each repository, we construct an isolated sandbox whose dependencies are inferred both by statically analyzing the repository's source code and by parsing its declared dependency manifests. This sandbox is preserved to ensure reproducible execution during subsequent seed extraction and task generation.

In this stage, we obtain \textbf{7,110} repositories. Their detailed distribution is reported in Figure~\ref{fig:discipline-pie}. Given the particularly active development of scientific agents in biology and biomedicine
\citep{mitchener2025bixbench,li2026genebench,anthropic2026biomysterybench},
we intentionally oversample biological repositories to better cover this important application domain. Prompts for keyword expansion, scientific-file detection, and scientific-field judgment are provided in Appendix~\ref{appendix:prompt}.

\subsubsection{Task Synthesis}
Given the repository pool, we develop a seed-program-centric method (\textbf{Panel (b) of Figure~\ref{fig:method}}) to deeply mine each repository and synthesize scientific tasks from its executable workflows. This method operates in two steps: we first extract executable seed programs from the raw repositories, and then derive diverse tasks from their underlying scientific workflows.

\paragraph{Seed Program Extraction.}
For each repository, we use DeepSeek-V4-Flash~\citep{deepseekai2026deepseekv4} to group related code files into one or more cohesive sets, each representing a complete scientific workflow (e.g. data denoise, computational analysis, and result generation). To standardize execution, DeepSeek-V4-Flash consolidates and rewrites each file set into a single entry-point code script, strictly preserving the original scientific logic. We then execute every candidate script within its sandbox; those that fail to run or produce errors are discarded, while successfully executed scripts are retained as \textbf{seed programs}.


\paragraph{Task Generation.}
Given a seed program, we use DeepSeek-V4-Flash to extract its underlying \textbf{scientific workflow} and represent it as a directed graph, where nodes denote data artifacts or computational operations and edges encode data dependencies. From this graph, we derive four complementary task types that target important capabilities in scientific discovery:
\begin{itemize}[leftmargin=*]
    \item \textbf{Reproduction} is constructed based on the inputs and outputs of the full workflow. It requires reproducing the entire analysis, targeting end-to-end workflow planning and execution.
    \item \textbf{Subflow} is constructed by extracting a specific part of the workflow. It requires executing this selected connected subgraph, targeting local dependency understanding and modular execution.
    \item \textbf{Completion} is constructed by removing a functional block from the complete code. It requires reconstructing the missing computational step from its surrounding context and dependencies, targeting dependency-aware workflow completion.
    \item \textbf{Correction} is constructed by injecting a controlled error into the program. It requires correcting this executable but scientifically invalid operation, targeting scientific error diagnosis and repair.
\end{itemize}
The construction logic for each task type is illustrated in the lower portion of Figure~\ref{fig:method}(b). For each task, we isolate the relevant segment of the seed program to construct a task-specific \textbf{reference program}. Executing this program yields the \textbf{reference answer}, which we then use to automatically generate an \textbf{evaluation script} tailored to the answer's format. Examples of each task type are provided in Appendix~\ref{appendix:data}.

\subsubsection{Verification}

We apply a multi-stage quality-verification pipeline (\textbf{Panel (c) of Figure~\ref{fig:method}}) to assess two complementary properties of each environment: \textbf{environment reliability}, which checks that the environment is complete, executable, reproducible, and reliably gradable; and \textbf{scientific validity}, which checks that the task is scientifically meaningful, correctly implemented, and internally consistent. Detailed criteria for both properties are provided in Appendix~\ref{appendix:verification}.

Each candidate environment is first evaluated by a GPT-5.5
\textbf{validator} according to these criteria. To recover valuable environments that initially fail
verification, we introduce \textbf{recursive improvement}. The validator
returns its error diagnosis and revision suggestions to the DeepSeek-V4-Flash
\textbf{generator}, which revises the environment before submitting it to the
validator again. This feedback loop between the generator and validator runs for up to three
rounds, and an environment is retained if it passes validation within this
budget. Finally, the validator assigns each retained environment a difficulty
score from 1 to 4 based on the required domain knowledge and the depth of the
analysis.

To further assess the quality of the automatically verified environments, we
randomly sample 600 tasks for independent manual review by three PhD
researchers. A task is accepted only if all three reviewers approve it,
yielding an acceptance rate of 95.5\% (Details are in Appendix~\ref{app:human-audit}. After automated filtering,
\pearl{} contains \textbf{97{,}659} training environments. Their distributions
across difficulty levels and task types are reported in
Appendix~\ref{app:data-distribution}. The prompts used for quality
verification, environment revision, and difficulty annotation are provided in
Appendix~\ref{appendix:prompt}.

%% file: sections/04_experiment.tex
\section{Experiment}
\subsection{Experimental Setup}

\paragraph{Agent Framework.}
We build our agent framework upon \textbf{mini-swe-agent}~\citep{yang2024sweagent}. Each task runs in an isolated sandbox with external network access disabled, and the agent interacts with the environment exclusively through a command-line interface. We impose a wall-clock time limit of 10{,}800 seconds per task and an execution timeout of 1{,}800 seconds per command, with a maximum of 100 agent iterations per task.

\input{tables/benchmark_overview}

\paragraph{Training Details.}
We use Qwen3.5-9B, Qwen3.5-27B, and Qwen3.5-35B-A3B as our backbone models.
Using DeepSeek-V4-Pro as the teacher model, we collect 64,477 successful
trajectories through rejection sampling from a pool of 97,659 training
environments. To reduce benchmark contamination, before trajectory
collection we removed environments whose source repositories matched,
or were near-duplicates of, repositories underlying the evaluation
benchmarks.  We fine-tune the
backbone models with a learning rate of
5.0$\times$10$^{-6}$, a batch size of 128, a maximum context length of
65,536 tokens, and a limit of 50 agent iterations per trajectory.

\paragraph{Evaluation Benchmarks.}
To thoroughly evaluate the scientific discovery capabilities of models trained on \pearl{} and other baselines, we assemble a suite of 6 scientific discovery benchmarks spanning ten scientific disciplines. Three of these are well-specified benchmarks with objectively verifiable outcomes: SciCode~\citep{tian2024scicode}, BioCoder~\citep{tang2023biocoder}, and MedAgentBench~\citep{jiang2025medagentbench}. The remaining three benchmarks, namely BixBench~\citep{mitchener2025bixbench}, BioMysteryBench~\citep{anthropic2026biomysterybench}, and ScienceAgentBench~\citep{chen2025scienceagentbench}, focus on open-ended scientific analyses. To ensure a fair and reliable evaluation, we exclude problems that are ill-posed or require languages outside our harness.  Table~\ref{tab:scientific-agent-benchmarks} summarizes the scale, disciplinary coverage, task format, and evaluation protocol of each benchmark. Detailed exclusion rules and further benchmark descriptions are provided in Appendix~\ref{app:benchmark-details}.

\paragraph{Models.}
We evaluate two groups of models. The first group comprises \textbf{frontier models}, including Claude-Opus-4.8~\citep{anthropic2026claudeopus48}, Gemini-3.6-Flash~\citep{googledeepmind2026gemini36flash}, GPT-5.5~\citep{openai2026gpt55}, Qwen3.8-Max~\citep{qwen38}, DeepSeek-V4-Pro~\citep{deepseekai2026deepseekv4highlyefficientmilliontoken}, GLM-5.2~\citep{glm5team2026glm5vibecodingagentic}, and Hy4-Preview~\citep{tencent2026hy4preview}. The second group comprises \textbf{baseline models}: the Qwen3.5 family~\citep{qwen3.5} (9B, 27B, and 35B-A3B). For all models, we set the temperature to 0.7, the maximum number of generated tokens to 32{,}768, and the reasoning effort to high.

\input{tables/main_results}

\subsection{Main Results}

We first evaluate the effectiveness of \pearl{} by comparing Qwen3.5 models of different sizes before and after supervised fine-tuning, together with several frontier models. The results are shown in Table~\ref{tab:main_results}. Overall, \textbf{training on \pearl{} consistently improves scientific-discovery performance across model scales and scientific domains}, covering both well-specified tasks and more challenging open-ended workflows. The gains are particularly pronounced for larger models, suggesting that stronger base models can better exploit the diverse and complex supervision provided by \pearl{}. Notably, Qwen3.5-27B improves by 10.4 percentage points on BioCoder and by 15.7 percentage points on ScienceAgentBench.

More importantly, \textbf{\pearl{} substantially narrows the gap between open models and frontier models}. After SFT, Qwen3.5-27B reaches 67.5\% on BioCoder, matching GLM-5.2, while Qwen3.5-35B-A3B achieves 20.0\% on SciCode (main), matching GPT-5.5. The 27B model also reaches 79.0\% on MedAgentBench and 69.8\% on BixBench-MCQ, outperforming Hy4-Preview (69.0\%) and GLM-5.2 (69.3\%), respectively, despite its substantially smaller model scale. These results demonstrate that \textbf{high-quality scientific-agent training data can yield large gains in scientific-discovery capabilities and enable relatively compact models to approach, or even surpass, frontier models on individual benchmarks}.

\paragraph{Comparing \pearl{} with Other Scientific-Discovery Data.}
We next compare \pearl{} with D3-Gym~\citep{moussa2026d3}, another method for generating scientific-agent training environments from code repositories. For a more comprehensive comparison, we also include MedAgentGym~\citep{xu2025medagentgym}, a collection of training environments for biomedical agent. D3-Gym and MedAgentGym provide 2{,}153 and 2{,}137 training trajectories, respectively. To compare under a similar data budget, we randomly subsample 2{,}000 trajectories from \pearl{} and fine-tune the same Qwen3.5-27B backbone using an identical training recipe (Table~\ref{tab:fair_comparison_27b}).
\input{tables/solvability}
Both baselines yield selective gains but also degrade performance on several
benchmarks. \textbf{MedAgentGym improves four of the eight settings} but decreases
BixBench (MCQ) by 8.1 percentage points, while \textbf{D3-Gym improves only three
settings} and reduces BixBench (Open) by 8.2 points and BioMystery by 7.8
points. These inconsistent transfer patterns are consistent with their limited
disciplinary coverage and limited exploitation of the scientific resources from
which their training data are constructed.
By contrast, \textbf{\pearl{} improves seven of the eight settings}, with gains ranging
from 0.7 to 5.2 percentage points and an overall average improvement of
2.1 points, despite using fewer than 10\% of the full training data.
The consistent gains at this matched data budget suggest that \textbf{\pearl{}'s
benefit arises from the diversity of its scientific disciplines, repositories,
and workflow-derived tasks, rather than from data scale alone.}

\input{tables/fair_comparison_27b}

\subsection{Ablation Study}

We conduct three ablation studies to examine how source diversity, environment
difficulty, and workflow-derived task composition contribute to
\pearl{}'s effectiveness. All ablation models are
trained by fine-tuning Qwen3.5-35B-A3B under the same training recipe and
evaluated on four representative benchmarks: BioCoder, MedAgentBench,
BixBench-MCQ, and ScienceAgentBench.

\paragraph{Scaling Repository Diversity.}

\begin{figure}[t]
\centering
\includegraphics[width=\linewidth]{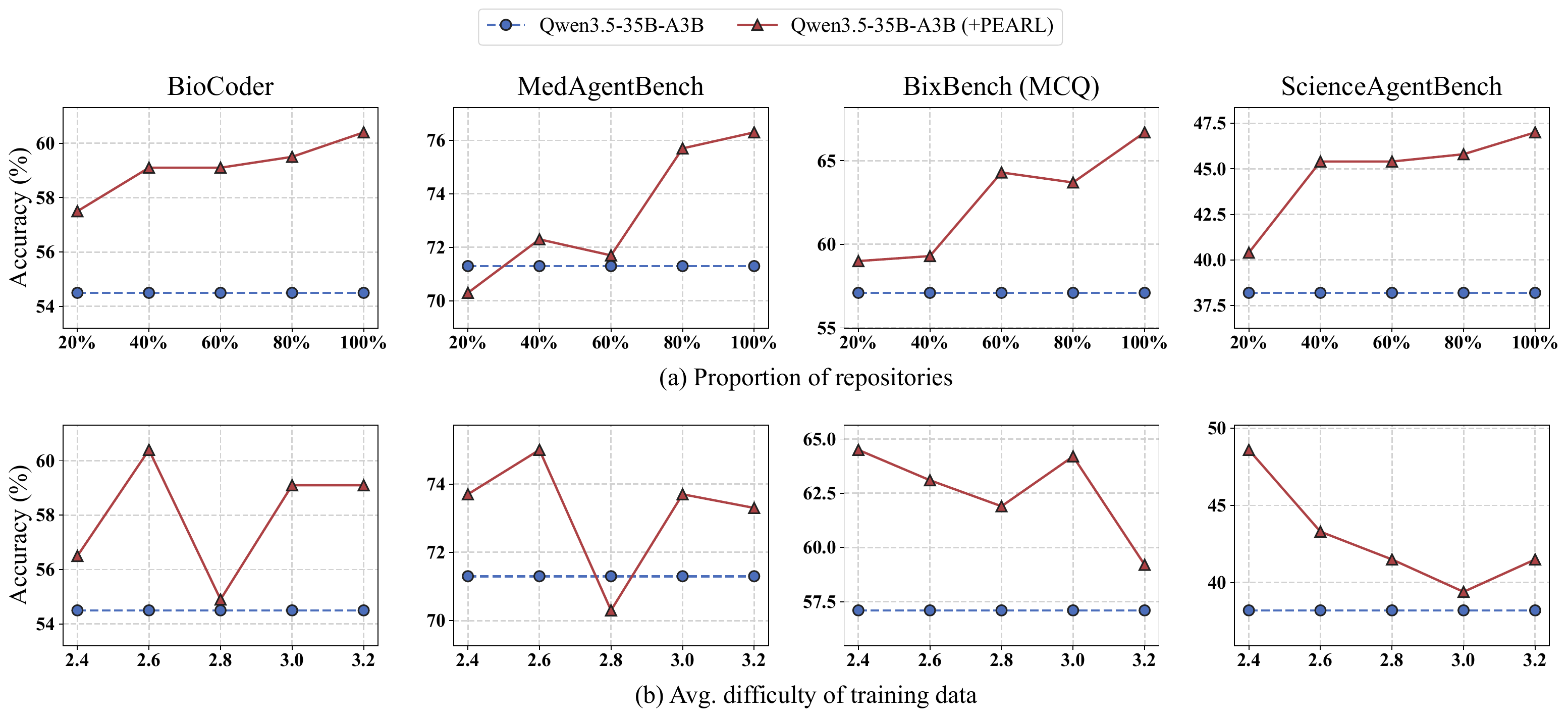}
\caption{
\textbf{Scaling repository diversity and environment difficulty under a fixed trajectory budget.}
The top row varies repository coverage, whereas the bottom row varies the
average difficulty of the training environments. The dashed line marks the
untrained Qwen3.5-35B-A3B backbone.
}
\label{fig:data_scaling}
\end{figure}

To isolate the effect of repository diversity, we fix the total number of
training trajectories, the average environment difficulty, and the proportions
of the four task types. We then vary the proportion of source repositories
from 20\% to 100\%. The top row of Figure~\ref{fig:data_scaling} shows that
\textbf{increasing repository diversity consistently improves training
effectiveness under a fixed trajectory budget}. Across all four benchmarks, performance exhibits an overall upward trend as repository coverage increases, despite minor fluctuations at intermediate points. This effect is particularly clear on MedAgentBench and BixBench (MCQ): at 20\% coverage, the trained model performs close to or even below the untrained backbone, whereas broader repository coverage eventually yields a clear improvement over the baseline. These results demonstrate \textbf{the value of expanding the breadth of scientific sources}, even with a fixed amount of training data.

\paragraph{Scaling Environment Difficulty.}

We next investigate how environment difficulty relates to task solvability and
training effectiveness. We first randomly sample 100 environments from each of
the four difficulty levels, D1--D4, and evaluate DeepSeek-V4-Pro on their
tasks. As shown in Table~\ref{tab:difficulty}, the success rate decreases from
94.0\% for D1 to 41.0\% for D4, while the average solution length increases
from 6.2 to 16.4 steps. These results indicate that the annotated difficulty
levels are consistent with empirical task difficulty.

We then construct five training subsets, each containing 30,000 trajectories,
with average environment difficulty scores of 2.4, 2.6, 2.8, 3.0, and 3.2,
respectively. We keep the number of repositories, the total trajectory budget,
and the proportions of the four task types fixed across subsets. As shown in
the bottom row of Figure~\ref{fig:data_scaling}, downstream performance
exhibits no consistent trend as the difficulty of the training environments
increases. Overall, \textbf{environment difficulty correlates with task
solvability but does not affect its effectiveness as training
data within the evaluated range}.

\paragraph{Removing Individual Task Types.}

We ablate each of the four workflow-derived task types in turn, yielding four
reduced datasets---w/o Reproduction, w/o Subflow, w/o Completion, and w/o Correction---
while controlling the total training-data size and source-repository
distribution. Table~\ref{tab:ablation_tasktype} reports the results.
\textbf{Removing any single task type degrades performance on every benchmark},
confirming that each contributes non-redundant supervision. This result
demonstrates the value of \textbf{mining multiple complementary forms of
supervision from each scientific workflow}.

\input{tables/ablation_tasktype}



%% file: tables/benchmark_overview.tex
\begin{table*}[t]
\centering

\captionsetup{
    width=\textwidth,
    justification=raggedright,
    singlelinecheck=false,
    skip=0pt
}

\caption{
    \textbf{Comparison of scientific agent benchmarks.}
    We summarize benchmark scale, covered disciplines, task format,
    and evaluation protocol.
}
\label{tab:scientific-agent-benchmarks}

\footnotesize
\renewcommand{\arraystretch}{1.15}
\setlength{\tabcolsep}{4pt}

\begin{threeparttable}

\begin{tabularx}{\textwidth}{
    @{}
    >{\raggedright\arraybackslash}p{0.17\textwidth}
    >{\centering\arraybackslash}p{0.06\textwidth}
    @{\hspace{8pt}}
    >{\raggedright\arraybackslash}X
    >{\centering\arraybackslash}p{0.16\textwidth}
    >{\centering\arraybackslash}p{0.17\textwidth}
    @{}
}

\specialrule{1.3pt}{0pt}{0pt}

\textbf{Benchmark}
&
\textbf{Size}
&
\textbf{Disciplines}
&
\textbf{\mbox{Open-ended}}
&
\textbf{Evaluation}
\\

\midrule

SciCode
&
65
&
Phys., Math., Mat., Chem., Bio.
&
\no
&
Rule-based
\\

BioCoder
&
154
&
Bioinfo.
&
\no
&
Rule-based
\\

MedAgentBench
&
300
&
Clin.\~Med.
&
\no
&
Rule-based
\\

BixBench
&
205
&
Bioinfo.
&
\yes
&
\mbox{LLM-as-Judge}
\\

BioMysteryBench
&
90
&
Bioinfo.
&
\yes
&
\mbox{LLM-as-Judge}
\\

ScienceAgentBench
&
102
&
Psych., Bioinfo., GIS, Comp.\~Chem.
&
\yes
&
\mbox{LLM-as-Judge}
\\

\specialrule{1.3pt}{0pt}{0pt}

\end{tabularx}

\vspace{2pt}

\parbox[t]{\textwidth}{
    \footnotesize
    \raggedright
    Abbreviations: Phys.\ (Physics), Math.\ (Mathematics),
    Mat.\ (Materials Science), Chem.\ (Chemistry), Bio.\ (Biology),
    Bioinfo.\ (Bioinformatics), Clin.\~Med.\ (Clinical Medicine),
    Psych.\ (Psychology and Cognitive Science), GIS (Geographic Information
    Science), Comp.\~Chem.\ (Computational Chemistry).
}

\end{threeparttable}

\end{table*}

%% file: tables/main_results.tex
\begin{table*}[t]
\centering

\caption{
    \textbf{Overall comparison across scientific discovery benchmarks.}
    MA-bench and SA-bench denote MedAgentBench and ScienceAgentBench, respectively.
    Scores are success rates (\%), and Avg is the mean over all benchmark columns.
    Within each size group (9B, A3B, 27B), \textbf{\textsc{\method}} is trained
    on the corresponding Qwen3.5 backbone; 
}

\label{tab:main_results}

\renewcommand{\arraystretch}{0.95}
\setlength{\tabcolsep}{4.5pt}

\resizebox{\textwidth}{!}{%
\fontsize{7.5}{9}\selectfont

\begin{tabular}{@{}lccccccccc@{}}
\toprule

\multirow{2}{*}{\textbf{Model}}
& \multirow{2}{*}{\textbf{BioCoder}}
& \multicolumn{2}{c}{\textbf{SciCode}}
& \multirow{2}{*}{\textbf{MA-Bench}}
& \multirow{2}{*}{\textbf{SA-Bench}}
& \multicolumn{2}{c}{\textbf{BixBench}}
& \multirow{2}{*}{\textbf{BioMystery}}
& \multirow{2}{*}{\textbf{Avg}}
\\

\cmidrule(lr){3-4}
\cmidrule(lr){7-8}

&
&
\textbf{Main}
&
\textbf{Sub}
&
&
&
\textbf{MCQ}
&
\textbf{Open}
&
&
\\

\midrule

\rowcolor{frontiercolor}
\multicolumn{10}{l}{\textit{\textbf{Frontier Models}}}
\\

\midrule

\logo{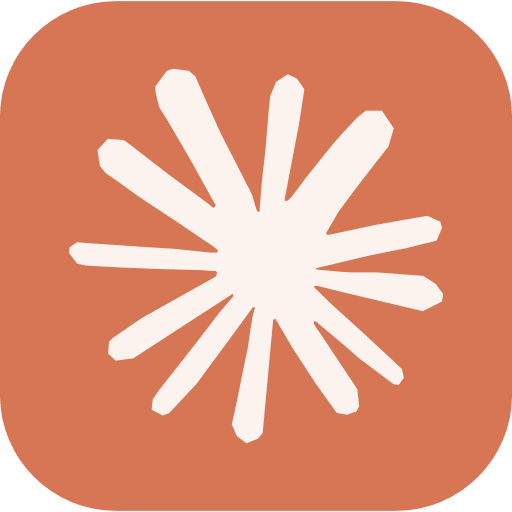}Claude-Opus-4.8
& 71.4
& 20.0 & 53.5
& 84.9
& 54.1
& 72.2 & 50.2
& 65.1
& 58.9
\\

\logo{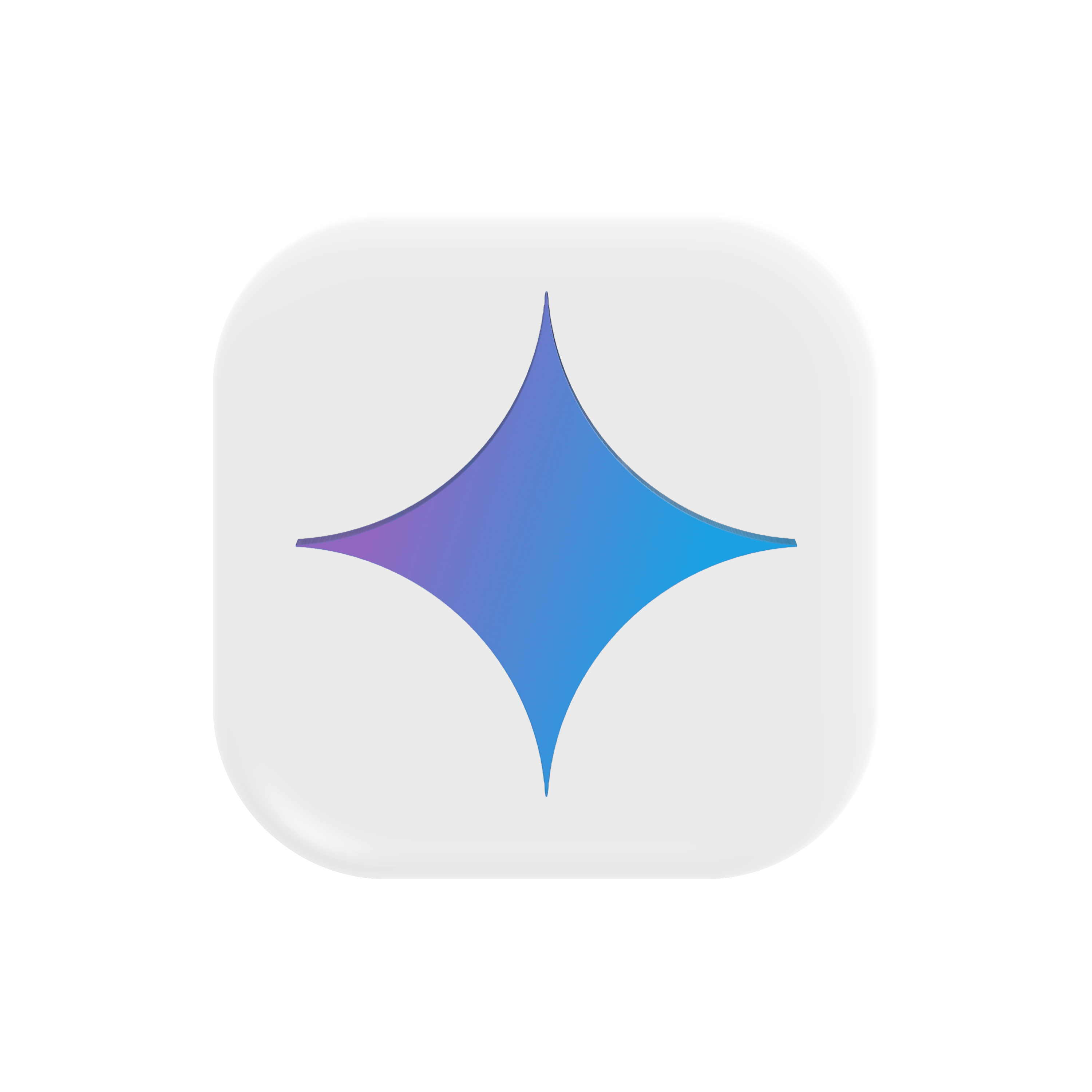}Gemini-3.6-Flash
& 70.8
& 20.0 & 59.0
& 92.3
& 55.9
& 73.7 & 47.8
& 56.7
& 59.5
\\

\logo{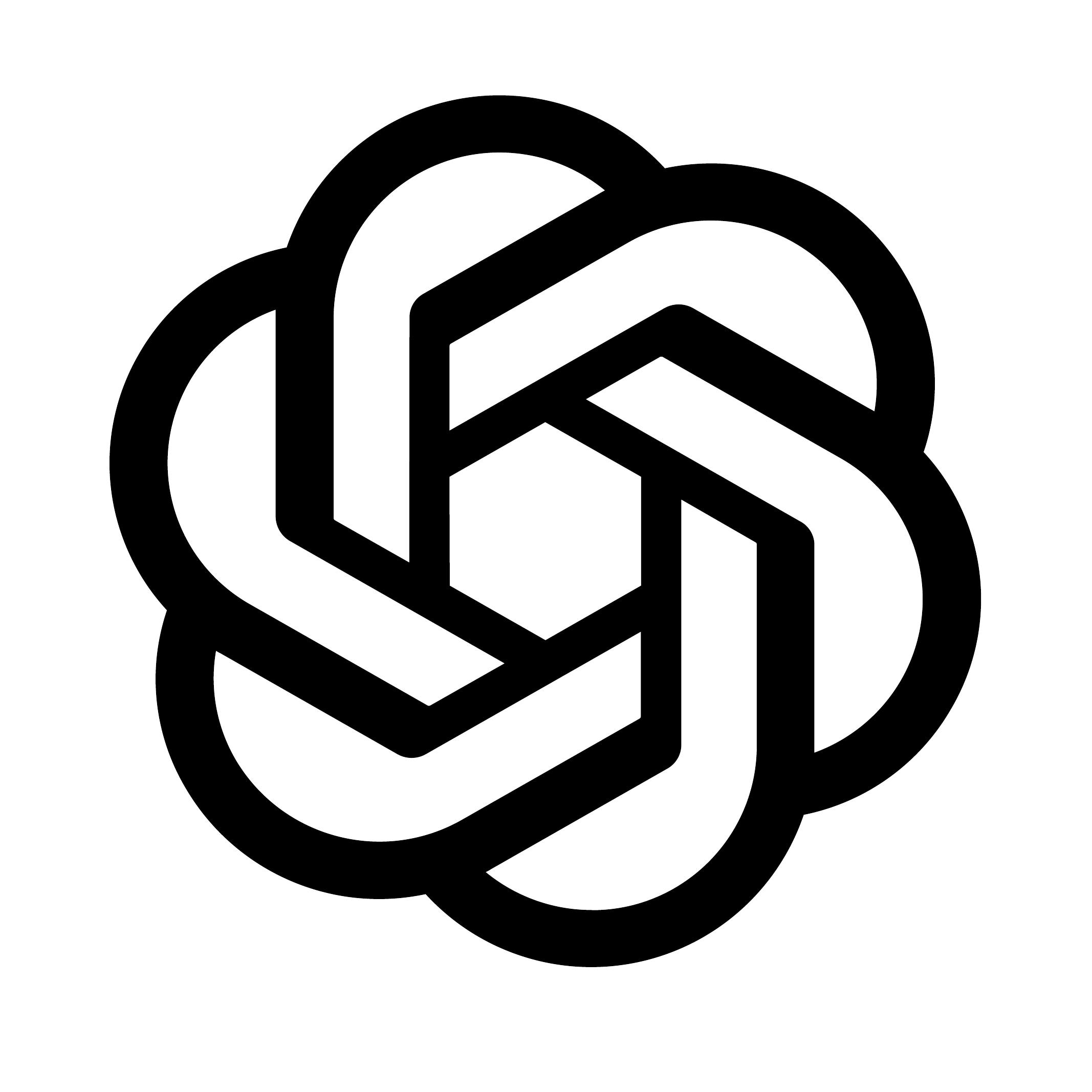}GPT-5.5
& 75.3
& 20.0 & 59.0
& 94.3
& 52.9
& 74.1 & 49.8
& 55.6
& 60.1
\\

\logo{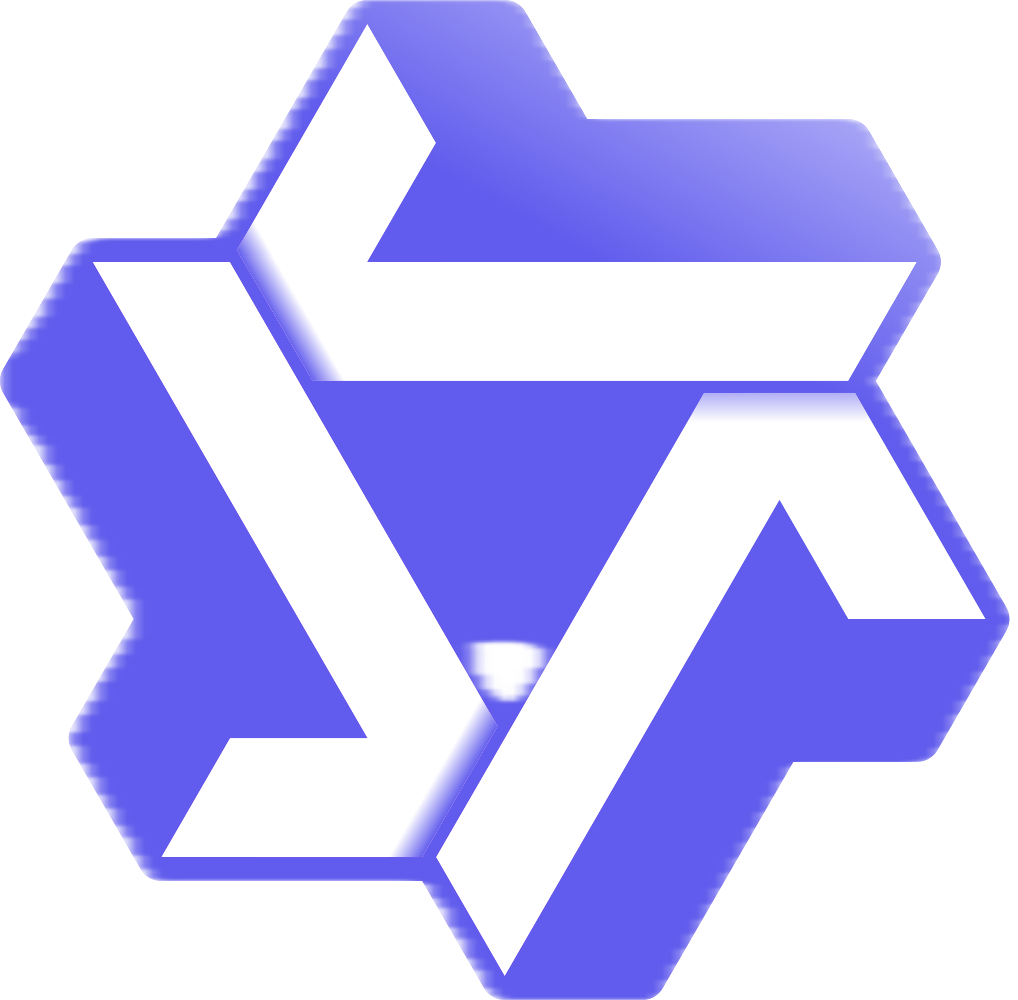}Qwen3.8-Max
& 77.9
& 23.1 & 57.6
& 86.3
& 64.7
& 77.6 & 45.9
& 64.4
& 62.2
\\

\logo{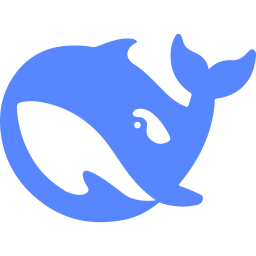}DeepSeek-V4-Pro
& 70.8
& 24.6 & 59.7
& 85.0
& 57.8
& 72.7 & 47.8
& 64.4
& 60.4
\\

\logo{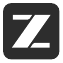}GLM-5.2
& 67.5
& 18.5 & 29.2
& 88.3
& 58.8
& 69.3 & 39.5
& 64.4
& 54.4
\\

\logo{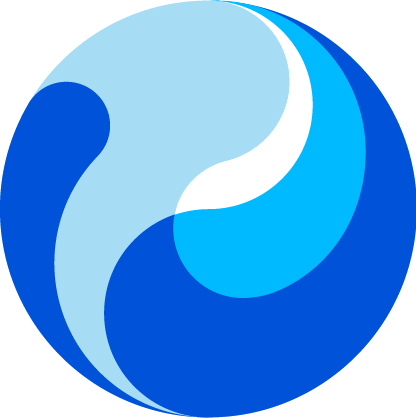}Hy4-Preview
& 80.5
& 18.5 & 54.9
& 69.0
& 61.8
& 72.2 & 49.8
& 44.4
& 56.4
\\

\midrule

\rowcolor{baseline9color}
\multicolumn{10}{l}{\textit{\textbf{9B Models}}}
\\

\midrule

\logo{qwen.png}Qwen3.5-9B
& 46.8
& 3.1 & 30.2
& 59.0
& 26.5
& 48.3 & 35.1
& 16.7
& 33.2
\\

\textbf{\quad +\,\textsc{\method}}
& \textbf{51.9}~\gain{5.1}
& \textbf{9.2}~\gain{6.1}
& \textbf{34.0}~\gain{3.8}
& \textbf{63.0}~\gain{4.0}
& \textbf{33.3}~\gain{6.8}
& \textbf{50.7}~\gain{2.4}
& \textbf{37.6}~\gain{2.5}
& \textbf{20.0}~\gain{3.3}
& \textbf{37.5}~\gain{4.3}
\\

\midrule

\rowcolor{ourscolor}
\multicolumn{10}{l}{\textit{\textbf{A3B Models}}}
\\

\midrule

\logo{qwen.png}Qwen3.5-35B-A3B
& 54.5
& 13.8 & 37.5
& 71.3
& 38.2
& 57.1 & 32.2
& 27.8
& 41.6
\\

\textbf{\quad +\,\textsc{\method}}
& \textbf{60.4}~\gain{5.9}
& \textbf{20.0}~\gain{6.2}
& \textbf{45.1}~\gain{7.6}
& \textbf{78.3}~\gain{7.0}
& \textbf{47.1}~\gain{8.9}
& \textbf{69.3}~\gain{12.2}
& \textbf{37.1}~\gain{4.9}
& \textbf{28.9}~\gain{1.1}
& \textbf{48.3}~\gain{6.7}
\\

\midrule

\rowcolor{baseline27color}
\multicolumn{10}{l}{\textit{\textbf{27B Models}}}
\\

\midrule

\logo{qwen.png}Qwen3.5-27B
& 57.1
& 12.3 & 42.7
& 75.7
& 34.3
& 61.5 & 34.1
& 36.7
& 44.3
\\

\textbf{\quad +\,\textsc{\method}}
& \textbf{67.5}~\gain{10.4}
& \textbf{16.9}~\gain{4.6}
& \textbf{50.3}~\gain{7.6}
& \textbf{79.0}~\gain{3.3}
& \textbf{50.0}~\gain{15.7}
& \textbf{69.8}~\gain{8.3}
& \textbf{39.5}~\gain{5.4}
& \textbf{38.9}~\gain{2.2}
& \textbf{51.5}~\gain{7.2}
\\

\bottomrule
\end{tabular}
}

\end{table*}

%% file: tables/solvability.tex
\begin{wraptable}{r}{0.4\textwidth}
\vspace{-8pt}
\centering
\small
\caption{Solvability across environment difficulty levels}
\label{tab:difficulty}
\renewcommand{\arraystretch}{1.1}
\setlength{\tabcolsep}{6pt}
\begin{tabular}{lcc}
\toprule
\textbf{Difficulty}
& \textbf{Success (\%)}
& \textbf{Avg. Steps} \\
\midrule
D1 & 94.0 & 6.2 \\
D2 & 86.0 & 8.1 \\
D3 & 57.0 & 15.1 \\
D4 & 41.0 & 16.4 \\
\bottomrule
\end{tabular}
\vspace{-8pt}
\end{wraptable}

%% file: tables/fair_comparison_27b.tex
\begin{table*}[t]
\centering
\caption{
\textbf{Data-scale-controlled comparison against scientific-agent training
baselines.} We subsample \textbf{2{,}000} training trajectories from
\textsc{\method} to obtain a data budget comparable to baselines.
}
\label{tab:fair_comparison_27b}

\setlength{\tabcolsep}{3pt}
\renewcommand{\arraystretch}{1.15}

\resizebox{\textwidth}{!}{%
\begin{tabular}{@{}l r c c c c c c c c c@{}}
\toprule
\multirow{2}{*}{\textbf{Model}}
& \multirow{2}{*}{\textbf{\# Traj.}}
& \multirow{2}{*}{\textbf{BioCoder}}
& \multicolumn{2}{c}{\textbf{SciCode}}
& \multirow{2}{*}{\textbf{MA-Bench}}
& \multirow{2}{*}{\textbf{SA-Bench}}
& \multicolumn{2}{c}{\textbf{BixBench}}
& \multirow{2}{*}{\textbf{BioMystery}}
& \multirow{2}{*}{\textbf{Avg}}
\\
\cmidrule(lr){4-5}
\cmidrule(lr){8-9}
& & & \textbf{Main} & \textbf{Sub}
& & & \textbf{MCQ} & \textbf{Open} & &
\\
\midrule

\logo{qwen.png}Qwen3.5-27B
& --
& 57.1
& 12.3
& 42.7
& 75.7
& 34.3
& 61.5
& 34.1
& \textbf{36.7}
& 44.3
\\
\midrule

\quad +\,\textsc{MedAgentGym}
& 2{,}137
& 60.4~\gain{3.3}
& \textbf{13.8}~\gain{1.5}
& 42.7
& \textbf{79.0}~\gain{3.3}
& 36.2~\gain{1.9}
& 53.4~\drop{8.1}
& 28.4~\drop{5.7}
& 32.2~\drop{4.5}
& 43.3~\drop{1.0}
\\

\quad +\,\textsc{D3-Gym}
& 2{,}153
& 59.1~\gain{2.0}
& 10.8~\drop{1.5}
& 38.9~\drop{3.8}
& 73.7~\drop{2.0}
& 36.2~\gain{1.9}
& 62.0~\gain{0.5}
& 25.9~\drop{8.2}
& 28.9~\drop{7.8}
& 41.9~\drop{2.4}
\\

\rowcolor{frontiercolor}
\textbf{\quad +\,\textsc{\method} (2k)}
& \textbf{2{,}000}
& \textbf{62.3}~\gain{5.2}
& \textbf{13.8}~\gain{1.5}
& \textbf{43.4}~\gain{0.7}
& 77.7~\gain{2.0}
& \textbf{38.1}~\gain{3.8}
& \textbf{63.9}~\gain{2.4}
& \textbf{37.6}~\gain{3.5}
& 34.4~\drop{2.3}
& \textbf{46.4}~\gain{2.1}
\\

\bottomrule
\end{tabular}%
}
\end{table*}

%% file: tables/ablation_tasktype.tex









\begin{table}[t]
\centering
\small
\caption{Ablation study on removing individual workflow-derived task types.}
\label{tab:ablation_tasktype}
\renewcommand{\arraystretch}{1.0}
\setlength{\tabcolsep}{5pt}

\resizebox{\linewidth}{!}{%
\begin{tabular}{@{}lccccc@{}}
\toprule
\textbf{Model Variant}
& \textbf{BioCoder}
& \textbf{MedAgentBench}
& \textbf{BixBench (MCQ)}
& \textbf{ScienceAgentBench}
& \textbf{Avg.} \\
\midrule

\rowcolor{frontiercolor}
\textbf{Qwen3.5-35B (+PEARL)}
& \textbf{60.4}
& \textbf{78.3}
& \textbf{69.3}
& \textbf{47.1}
& \textbf{63.8} \\
\midrule

w/o Reproduction
& 54.5~\drop{5.9}
& 70.7~\drop{7.6}
& 63.9~\drop{5.4}
& 44.1~\drop{3.0}
& 58.3~\drop{5.5} \\

w/o Subflow
& 55.8~\drop{4.6}
& 73.0~\drop{5.3}
& 60.5~\drop{8.8}
& 41.2~\drop{5.9}
& 57.6~\drop{6.2} \\

w/o Completion
& 59.7~\drop{0.7}
& 71.7~\drop{6.6}
& 62.0~\drop{7.3}
& 43.1~\drop{4.0}
& 59.1~\drop{4.7} \\

w/o Correction
& 55.2~\drop{5.2}
& 72.0~\drop{6.3}
& 62.0~\drop{7.3}
& 46.1~\drop{1.0}
& 58.8~\drop{5.0} \\

\bottomrule
\end{tabular}%
}
\end{table}

%% file: sections/05_conclusion.tex
\section{Conclusion}

We introduced \poseidon{}, an automated pipeline for generating diverse, verifiable training environments from scientific code repositories at scale. By deriving four complementary task types from executable workflows and refining failed candidates through recursive improvement, \poseidon{} produces richer supervision from existing scientific code. Using this pipeline, we constructed \pearl{}, a collection of 97,659 environments from 7,110 repositories across 10 scientific disciplines. Supervised fine-tuning on \pearl{} improves Qwen3.5 models across six scientific-agent benchmarks. Controlled ablations further show that repository diversity and complementary task types contribute to these gains, supporting the value of extracting supervision beyond end-to-end workflow execution. Together, \poseidon{} and \pearl{} provide a practical foundation for scaling training environments and developing more capable scientific discovery agents.

%% file: sections/A1_appendix_head.tex
\section{Data Distribution}
\label{app:data-distribution}

\begin{figure}[htbp]
    \centering
    \begin{minipage}[t]{0.48\textwidth}
        \centering
        \vspace{0pt}
        \includegraphics[height=3.8cm]{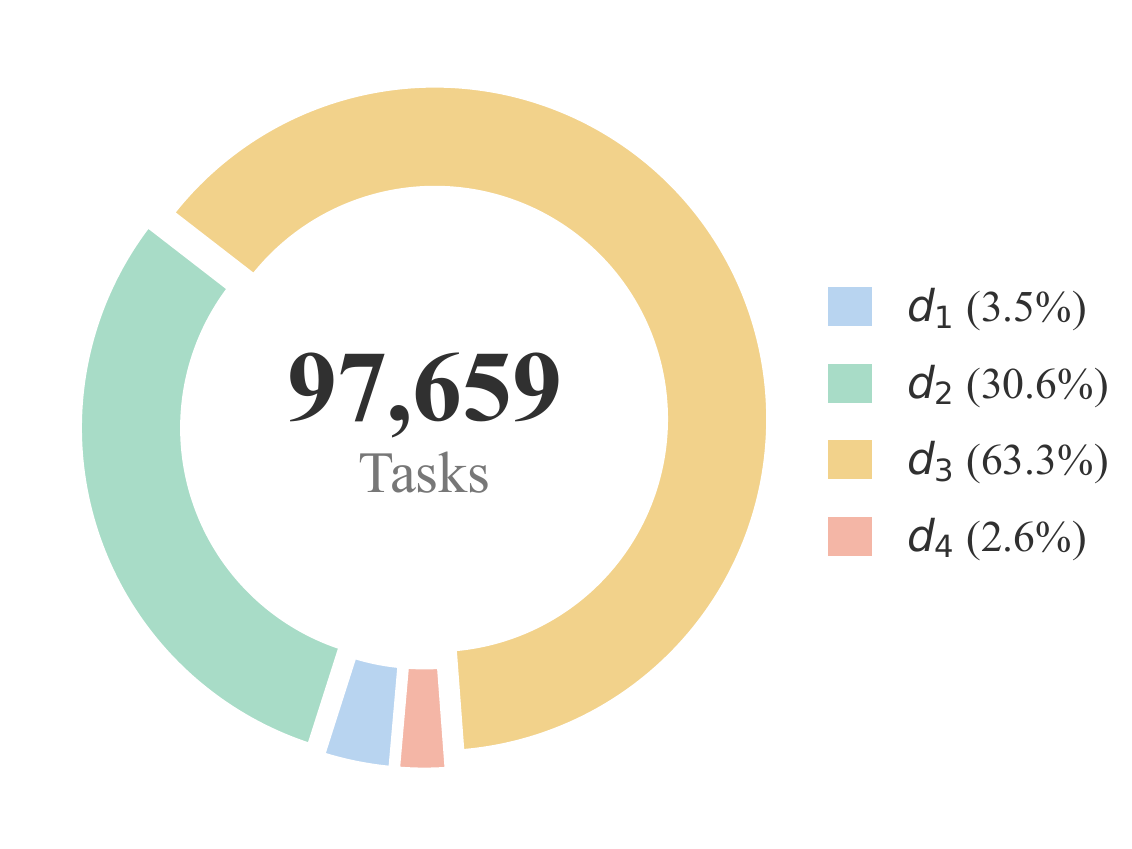}
        \caption{Difficulty distribution of the tasks.}
        \label{fig:distill-difficulty}
    \end{minipage}
    \hfill
    \begin{minipage}[t]{0.48\textwidth}
        \centering
        \vspace{0pt}
        \includegraphics[height=3.8cm]{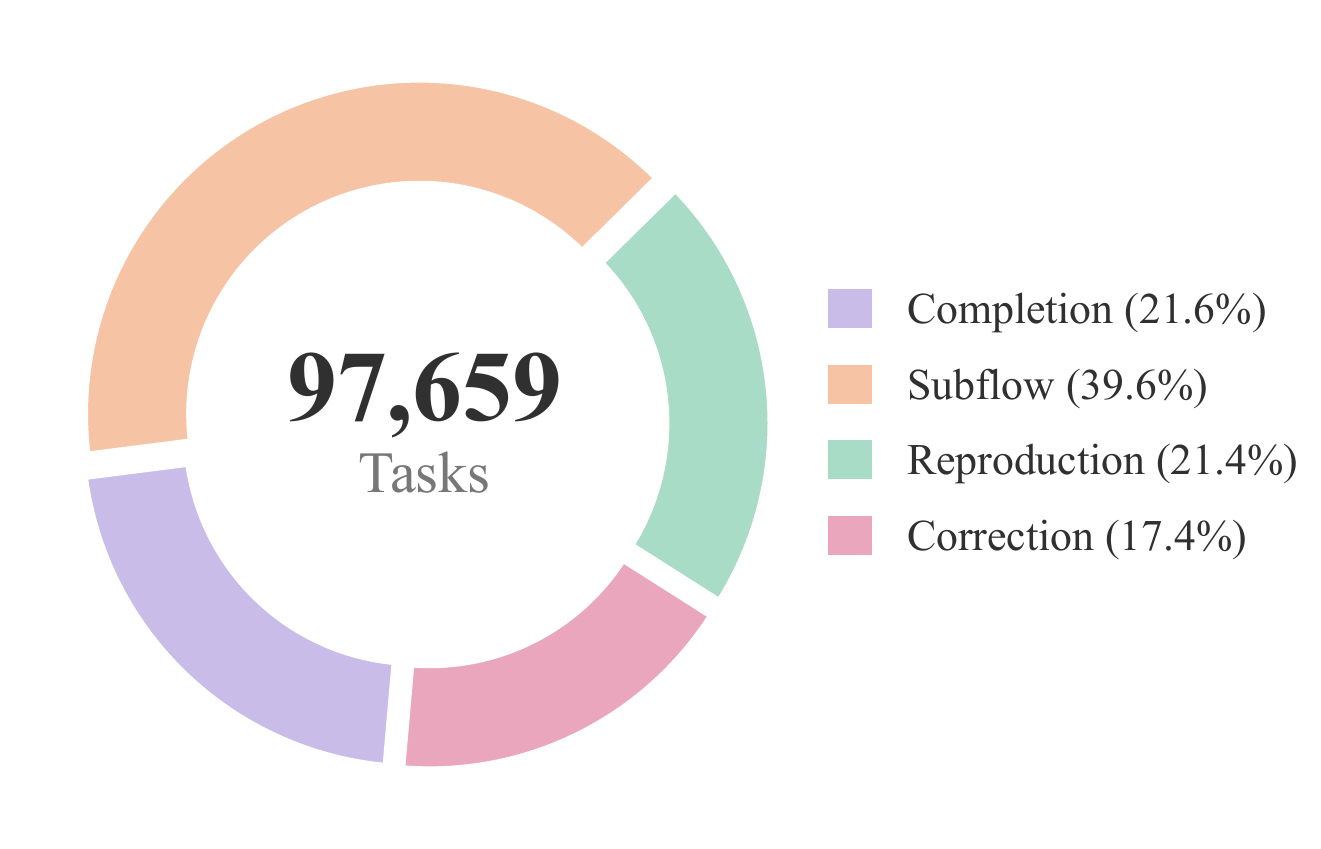}
        \caption{Task-type distribution of the tasks.}
        \label{fig:distill-tasktype}
    \end{minipage}
\end{figure}

Figure~\ref{fig:distill-difficulty} and
Figure~\ref{fig:distill-tasktype} show the difficulty and
task-type distributions of our \textbf{97,659} tasks.

%% file: sections/A1_5_verification.tex
\section{Verification Criteria Details}\label{appendix:verification}
\paragraph{Environment Reliability.}
We first verify that each environment is structurally complete, executable,
reproducible, and reliably gradable. Specifically, we assess \textbf{(1) data
integrity}, requiring all referenced data files to exist and their paths to be
valid and consistent with the package structure; \textbf{(2) executability},
requiring the reference program to run successfully within the task-specific
sandbox; \textbf{(3) answer reproducibility}, requiring execution of the
reference program to reproduce the reference answer exactly; and
\textbf{(4) gradability}, requiring the task instruction to unambiguously
specify the requested outputs and the evaluator to accept the reference answer
while rejecting representative incorrect answers.

\paragraph{Scientific Validity.}
We next verify that each task is scientifically meaningful, correctly
implemented, and internally consistent. Specifically, we assess
\textbf{(1) scientific correctness}, requiring the task to follow valid
scientific logic and define a meaningful computation; \textbf{(2) program
correctness}, requiring the reference program to faithfully implement the
computation specified in the instruction; and \textbf{(3) repair consistency},
requiring the intended defect in a correction task to be present in the supplied
code and the resulting behavior to match the failure described in the
instruction.

%% file: sections/A1_6_human_verification.tex
\section{Independent Human Audit of Environment Quality}
\label{app:human-audit}

The automatic verifier enables scalable quality control, but its judgments
alone do not establish the scientific validity of the generated
environments. We therefore conduct an independent human audit of the
retained environments.

\subsection{Sampling and Stratification}

We randomly sample 600 environments from the final \pearl{} collection
after automatic verification. Sampling is stratified by scientific
discipline, task type, and annotated difficulty level. We use proportional
allocation to preserve the distribution of the full collection, while
enforcing a minimum of 10 environments in every non-empty
discipline--task-type stratum. The audit set covers all scientific
disciplines, the four task types (Reproduction, Subflow, Completion, and
Correction), and all four difficulty levels whenever such combinations are
present in the collection.

The audited environments are disjoint from the examples used to develop or
tune the automatic verifier and from the examples included in the
verification prompts. Auditors do not observe the verifier's scores,
labels, revision history, or model-generated explanations.

\subsection{Auditors and Blinding}

Each environment is independently reviewed by three auditors. All auditors
hold a Ph.D. or equivalent research experience in a relevant scientific or
computational field. At least one auditor assigned to each environment has
domain expertise matching the environment's scientific discipline.

Auditors receive only the task instruction, input data, sandbox, reference
program, evaluation script, and documented execution requirements. During
the initial semantic review, they do not observe the automatic verifier's
decision or confidence score. The reference answer is hidden unless it is
required for a separate execution-consistency check.

Auditors follow a written rubric and submit their decisions independently.
Disagreements are discussed only after all independent decisions have been
recorded. Cases that remain unresolved are reviewed by a senior auditor,
whose adjudication is recorded separately.

\subsection{Audit Rubric}

For each environment, auditors assess the following criteria:

\begin{enumerate}
    \item \textbf{Scientific relevance}: the task addresses a meaningful
    scientific computation, analysis, simulation, or interpretation problem.

    \item \textbf{Task identifiability}: the requested outputs, units,
    thresholds, aggregation rules, and acceptance criteria are sufficiently
    specified to determine a valid answer.

    \item \textbf{Scientific correctness}: the stated computation follows
    scientifically valid logic under the assumptions given in the task.

    \item \textbf{Program correctness}: the reference program implements
    the computation specified by the task, including indexing, units,
    filtering, aggregation, and numerical conventions.

    \item \textbf{Artifact consistency}: the input files, reference
    program, reference answer, and evaluation script are mutually
    consistent.

    \item \textbf{Execution and reproducibility}: the reference program
    executes successfully in the provided sandbox and reproduces the
    reference output under the documented conditions.

    \item \textbf{Evaluation validity}: the evaluation script accepts the
    reference answer and rejects representative materially incorrect
    answers.
\end{enumerate}

For each criterion, auditors record one of three judgments:
\texttt{pass}, \texttt{fail}, or \texttt{uncertain}, together with a
short evidence-based rationale. An environment is considered fully valid
only if all criteria receive a \texttt{pass} judgment after adjudication.
An \texttt{uncertain} judgment is not counted as a pass.

For Completion and Correction tasks, auditors additionally verify that the
missing or faulty computation is present as described, that the reported
symptom is reproducible, and that the intended repair preserves the stated
scientific objective.

\subsection{Execution Checks}

Static inspection and execution evidence are recorded separately. For each
audited environment, an isolated execution procedure checks:

\begin{enumerate}
    \item successful completion without an unreported exception;
    \item existence and integrity of all required input files;
    \item reproduction of the reference output in two independent executions;
    \item acceptance of the reference output by the evaluation script; and
    \item rejection of at least two representative incorrect outputs,
    including a numerically or semantically plausible error when
    applicable.
\end{enumerate}

If execution cannot be completed because of an infrastructure failure, the
case is marked as \texttt{unverified} rather than being treated as
scientifically valid.

\subsection{Audit Outcome}

Among the 600 audited environments, 573 (95.5\%) satisfied all audit
criteria, including scientific relevance, task identifiability, scientific
correctness, program correctness, artifact consistency, execution
reproducibility, and evaluation validity. The remaining 27 environments
failed at least one criterion or could not be conclusively verified.

%% file: sections/A2_benchmark_details.tex
\section{Benchmark Details}
  \label{app:benchmark-details}

  \paragraph{SciCode.}
  SciCode~\citep{tian2024scicode} evaluates scientific code generation through 80 main problems decomposed into 338 subproblems. It covers physics, mathematics, materials science, chemistry, and biology. Solutions are
  evaluated using expert-written test cases with numerical-tolerance checks. We evaluate on SciCode's test split (65 main problems / 288 gradeable subproblems). The official 80/338 figure combines the test and dev splits; we report test only, which corresponds to the official \texttt{STEP\_NUM=288} denominator used when evaluating that split.

  \paragraph{BioCoder.}
  BioCoder~\citep{tang2023biocoder} evaluates the generation of functionally correct code derived from real-world bioinformatics repositories. We evaluate 154 of the 207 public problems: 50 Java problems are omitted by
  our Python-only harness, and three Python problems are excluded because their reference solutions do not execute under any tested library configuration. BioCoder uses differential testing against reference
  implementations; a reference-solution sweep passes all 154 retained problems.

  \paragraph{MedAgentBench.}
  MedAgentBench~\citep{jiang2025medagentbench} evaluates agents in a simulated electronic health record environment. It contains 300 tasks across ten clinical workflow categories, requiring agents to query and update
  patient records through FHIR interfaces. Performance is determined by rule-based verification of the resulting EHR state.

  \paragraph{BixBench.}
  BixBench~\citep{mitchener2025bixbench} evaluates open-ended bioinformatics analysis over realistic datasets and computational environments. Its 205 questions require agents to inspect data, execute multi-step analyses,
  and produce evidence-supported answers. Responses are evaluated using an LLM-as-Judge protocol.

  \paragraph{BioMysteryBench.}
  BioMysteryBench~\citep{anthropic2026biomysterybench} assesses agents on open-ended biological research problems involving modalities such as genomics, transcriptomics, proteomics, and metabolomics. Agents must analyze
  raw scientific data and report conclusions supported by computational evidence. We evaluate 90 tasks using an LLM-as-Judge protocol.

  \paragraph{ScienceAgentBench.}
  ScienceAgentBench~\citep{chen2025scienceagentbench} comprises 102 tasks derived from 44 scientific publications across psychology and cognitive science (28), bioinformatics (27), geographical information science (27), and computational chemistry (20). Each task requires an agent to implement and execute a scientific workflow. Grading is per-task: 38 tasks are decided purely by rule-based checks of the generated code and output artifacts, while the remaining 64 additionally invoke an LLM judge that scores a generated figure against a reference image at a threshold of 60. Following prior practice, we report both the full 102-task set and the 38-task rule-based subset.

%% file: sections/A3_data_examples.tex
\section{Details Demonstration}

\subsection{Data Example}\label{appendix:data}

\begin{detail}{Reproduction}{detail:MHV}
You are required to predict the human-infectivity probability of viral proteins based on their 6-amino-acid peptide content relative to a reference database.\\

\textbf{Problem to Solve}

Mouse hepatitis virus (MHV) is a mouse coronavirus; the animal-host reference is therefore the appropriate database. You are given the viral protein FASTA file \texttt{data/proteome\_and\_protein\_sequences/MHV.fasta} and the precomputed reference database \texttt{data/sphak/data/animal\_reference\_database.pkl}. The reference database stores, for each viral family, the 6-amino-acid peptide content of family members known to infect humans and of family members not known to infect humans, along with the corresponding total peptide counts.

For every protein sequence in \texttt{MHV.fasta}, determine the best matching viral family using the 6-amino-acid peptide content of the sequence relative to the family profiles. Define the best family as the family whose stored 6-amino-acid peptide set has the largest overlap with the query sequence's 6-amino-acid peptide set. Then estimate a human-infectivity probability for the sequence using a naive Bayes classifier with the following exact specification:

\begin{itemize}
    \item Use only 6-amino-acid peptides ($k=6$).
    \item For the assigned family, consider each 6-mer in the query sequence that is present in the family's peptide set (i.e., in the union of its host-positive and host-negative peptide keys).
    \item Apply position-based deduplication: a peptide contributes to the likelihood only if its positions (indices $i$ through $i+5$) are disjoint from the positions of all previously contributed peptides in the sequence (process peptides in order of their start index).
    \item For each contributing peptide, let $h$ be its count in the family's host-positive data and $nh$ its count in the family's host-negative data. Let \texttt{total\_homo} and \texttt{total\_non\_homo} be the family's total host-positive and host-negative peptide counts, respectively. Let \texttt{vocab\_size} $= 20^6$. Use Laplace smoothing with $\alpha = 0.1$, so $p_{\text{homo}} = (h + 0.1) / (\texttt{total\_homo} + 0.1 \times \texttt{vocab\_size})$ and $p_{\text{non}} = (nh + 0.1) / (\texttt{total\_non\_homo} + 0.1 \times \texttt{vocab\_size})$.
    \item Apply temperature scaling with $T = 1.5$: raise both $p_{\text{homo}}$ and $p_{\text{non}}$ to the power $1.5$, then normalize them so they sum to 1.
    \item Accumulate the natural log of each normalized probability over all contributing peptides. After processing all contributing peptides, divide the accumulated log-likelihoods by the number of unique contributing position-sets (i.e., the number of peptides that actually contributed).
    \item Use uniform priors ($0.5$ for human-positive, $0.5$ for human-negative). The posterior probability of human infectivity is computed as $\exp(\log(0.5) + \text{averaged\_log\_}p_{\text{homo}})$ divided by the sum of $\exp(\log(0.5) + \text{averaged\_log\_}p_{\text{homo}})$ and $\exp(\log(0.5) + \text{averaged\_log\_}p_{\text{non}})$, using the log-sum-exp trick for numerical stability. If no peptide contributes, set the posterior to $0.5$.
\end{itemize}

Additionally, compute the fraction of sequence positions covered by 6-mers that match the assigned family's peptide set (i.e., the number of positions that fall within at least one matching 6-mer, divided by the total sequence length). A sequence is predicted to infect humans only when the estimated human-infectivity probability is $> 0.5$ and the coverage fraction is $> 0.30$.\\

Report for each sequence: (a) best family label, (b) human-infectivity probability rounded to 4 decimal places, (c) coverage fraction rounded to 4 decimal places, (d) final binary prediction (1 = predicted human-infecting, 0 = not predicted human-infecting).

\end{detail}

\begin{detail}{Subflow}{detail:GMM}

You are required to estimate the per-line zero-mean two-component Gaussian-mixture uncertainty parameters for a 118-bus transmission system.\\

\textbf{Problem to Solve}

You are estimating the per-line zero-mean two-component Gaussian-mixture uncertainty parameters (the $\eta_l$ model) for the 118-bus transmission system.

The intermediate result (equivalent to the product of step 9 of the workflow) is provided: for generation seed 42, for each transmission line $l$ represented by a row of \texttt{data/H\_wind\_118\_matrix.csv}, the bivariate sample
\[
X_l = [\Omega_i, \Lambda_{l,i}], \quad i = 1\dots 8000
\]
is available, where $\Omega$ is the total wind deviation and $\Lambda_l$ is that line's flow deviation. If you need to materialize it, use the established synthetic-sample protocol: $N=8000$, $D=10$, scale $\gamma=0.01$, seed 42, and the sensitivity rows from \texttt{data/H\_wind\_118\_matrix.csv} in file order.

Continue from this intermediate result. The $\eta_l$ stage parameterizes each line by a two-component mixture of bivariate zero-mean Gaussian distributions, so do not replace this with a raw single-Gaussian covariance estimate of the pooled bivariate sample, and do not carry over a three-regime structure from an earlier univariate diagnostic. Component means are fixed at zero.

For each line, two covariance families are to be considered and compared by BIC on the fitted log-likelihood:
\begin{itemize}
    \item \texttt{spherical}: component covariance matrices are $\tau_k \times I$;
    \item \texttt{tied\_shape}: component covariance matrices are $\tau_k \times C$, where $C$ is a common $2\times2$ covariance shape shared by both components and $\tau_k$ is component-specific. \textbf{In this \texttt{tied\_shape} model, $C$ is fixed at the pooled sample covariance of the bivariate sample $X_l$ (normalized so that its trace equals the dimension 2), and only the weights and $\tau_k$ are estimated by EM.}
\end{itemize}

Use the BIC penalty convention for these bivariate models: the \texttt{spherical} model has 3 free parameters and the \texttt{tied\_shape} model has 5 free parameters, with $N=8000$ observations used in the $\log(N)$ penalty. For each line, choose the covariance family with the lower BIC, using 10 repeated fits per family with different random initializations (e.g., seeds 1000 through 1009) so that a clearly poor local optimum is not reported as the final answer.

Report three quantities, all for generation seed 42:
\begin{enumerate}
    \item[(a)] Among all line models in \texttt{data/H\_wind\_118\_matrix.csv}, the number of lines whose BIC-best covariance family is \texttt{tied\_shape}.
    \item[(b)] The BIC-best covariance family for the first line in file order: report \texttt{spherical} or \texttt{tied\_shape}.
    \item[(c)] For that same first line and for its component with the largest mixture weight, the covariance entry $\text{Cov}(\Omega, \Lambda_l)$ (the off-diagonal entry of that component's $2\times2$ covariance matrix).
\end{enumerate}

\end{detail}

\begin{detail}{Completion}{detail:GISAXS}

You are required to reimplement a missing scientific computation step that converts a detector image into reciprocal-space coordinates, and then count the predicted grating peaks within the measured window.\\

\textbf{Problem to Solve}

A GISAXS experiment was performed on a one-dimensional grating with a nominal period of 1390 nm. A single detector exposure was recorded with a sample-detector distance of $\text{DETX0} = 100.4\text{ mm}$, and the recorded sample azimuth is $\text{PHI} = -0.04^\circ$. The raw detector image and detector/scan metadata have already been loaded into the variables \texttt{image\_array} and \texttt{params\_dict\_list}. The scientific goal is to index the GISAXS pattern by converting the detector image into reciprocal space, applying the standard relative-intensity normalization, and overlaying the predicted grating peak positions.

The program below is incomplete: exactly one scientific computation step has been removed---the conversion of the detector image into the reciprocal-space coordinates \texttt{qx}, \texttt{qy}, \texttt{qz}. Reimplement that missing step only, using \texttt{DETX0} and the available metadata/raw image. Keep all other calls unchanged. Then compute the number of predicted grating peaks (grating period 1390 nm) that fall inside the measured \texttt{qy}-\texttt{qz} window, separately for $\text{PHI} = -0.04$ and $\text{PHI} = 5.0$. Report the two counts as integers.\\

Use the following exact conventions for the missing step:
\begin{itemize}
    \item Each detector pixel is treated as a 1D array of length \texttt{n\_points = len(image\_array[i])}.
    \item Pixel indices are centered: \texttt{pixel\_indices = np.arange(n\_points) - (n\_points - 1) / 2.0}.
    \item Detector coordinates in mm: \texttt{x\_det = pixel\_indices * pixel\_size}, \texttt{y\_det = pixel\_indices * pixel\_size}, where \texttt{pixel\_size = params\_dict\_list[i]['pixel\_size']} (in mm).
    \item Exit angles: \texttt{alpha\_f = np.arctan(y\_det / distance)}, \texttt{theta\_f = np.arctan(x\_det / distance)}, where \texttt{distance = params\_dict\_list[i]['distance']} (in mm).
    \item Incident angle: \texttt{alpha\_i = 0.0} (grazing incidence, simplified).
    \item Azimuth phi (in degrees) is taken from \texttt{params\_dict\_list[i]['phi']} and converted to radians: \texttt{phi\_rad = np.deg2rad(phi)}.
    \item Wavelength: \texttt{wavelength = params\_dict\_list[i]['wavelength']} (in \AA), and $k = 2\pi / \text{wavelength}$.
    \item Use the following exact formulas (small-angle approximation is NOT used; use the full trigonometric expressions):
    \begin{align*}
        q_x &= k \cdot (\cos(\alpha_f) \cos(\theta_f) - \cos(\alpha_i)) \\
        q_y &= k \cdot (\cos(\alpha_f) \sin(\theta_f) \cos(\phi_{\text{rad}}) + \sin(\alpha_f) \sin(\phi_{\text{rad}})) \\
        q_z &= k \cdot (\sin(\alpha_f) \cos(\phi_{\text{rad}}) - \cos(\alpha_f) \sin(\theta_f) \sin(\phi_{\text{rad}}))
    \end{align*}
    \item Store the results as numpy arrays \texttt{qx\_array}, \texttt{qy\_array}, \texttt{qz\_array} with shape \texttt{(n\_images, n\_points)}.
\end{itemize}

Note: The functions \texttt{calibration.get\_angle}, \texttt{calibration.get\_sr}, \texttt{calibration.get\_rel\_intensity}, and \texttt{gratings.calculate\_q} are provided by the environment (they may be inline replacements, not the original \texttt{xray\_scatter\_py} package). Do not modify those calls.

\begin{lstlisting}[language=Python]
import numpy as np
from xray_scatter_py import calibration, gratings

DETX0 = 100.4   # mm
B = 1390        # nm
PHI_1 = -0.04
PHI_2 = 5.0

# params_dict_list and image_array are already loaded from the exposure.

theta_array, azimuth_array = calibration.get_angle(DETX0, params_dict_list, image_array)

# ============ MISSING SCIENTIFIC STEP ============
# Compute qx_array, qy_array, qz_array, the reciprocal-space coordinates for
# every detector pixel, from DETX0, params_dict_list, and image_array.
# TODO: implement this step
# =================================================

# Solid-angle corrected relative intensity (do not modify)
sr_array = calibration.get_sr(DETX0, params_dict_list, theta_array)
image_array_rel = calibration.get_rel_intensity(params_dict_list, image_array, sr_array)

# Measured q window
xmin, xmax = np.min(qy_array), np.max(qy_array)
ymin, ymax = np.min(qz_array), np.max(qz_array)

def count_predicted_peaks(phi):
    qx, qy, qz, qx_0, qy_0, qz_0 = gratings.calculate_q(
        params_dict_list, image_array, phi=phi, b=B)
    inside = (qy >= xmin) & (qy <= xmax) & (qz >= ymin) & (qz <= ymax)
    return int(np.sum(inside))

n_peaks_neg0_04 = count_predicted_peaks(PHI_1)
n_peaks_5 = count_predicted_peaks(PHI_2)
print(n_peaks_neg0_04, n_peaks_5)
\end{lstlisting}

\end{detail}

\begin{detail}{Correction}{detail:NEGF}

You are required to diagnose and fix a scientific error in a finite-difference gradient checker for a bistable Poisson-NEGF system, and then verify the corrected audit.\\

\textbf{Problem to Solve}

A Poisson-NEGF conductance audit compares analytic adjoint gradients of the self-consistent conductance with central finite-difference estimates at three states of a normalized sweep: depressed ($s = -1$), middle ($s \approx +0.0667$), and potentiated ($s = +1$). The audit is accepted only when the worst relative error over parameters (\texttt{UB}, \texttt{Pr}, \texttt{Vread}) is $< 1\times10^{-5}$.

The current implementation, when run, shows the following behavior:
\begin{itemize}
    \item At $s = -1$ and $s \approx +0.0667$, the analytic adjoint and finite-difference gradients agree to well below the $1\times10^{-5}$ acceptance threshold.
    \item At $s = +1$ (potentiated), the audit fails: the finite-difference gradient is sharply inconsistent with the analytic adjoint for at least one parameter, \texttt{passed} is false, and the reported \texttt{worst\_rel\_err} is far above $1\times10^{-5}$.
    \item No exception is raised, and both perturbed SCF solves converge normally.
    \item It is known from continuation studies that at the potentiated state the self-consistent Poisson-NEGF fixed-point map is bistable, so the conductance-versus-parameter curve has more than one branch.
\end{itemize}

The finite-difference checker under suspicion is implemented as follows (the rest of the Poisson-NEGF/adjoint machinery is known to be correct):

\begin{lstlisting}[language=Python]
def fd_grad(G, theta, s_k, Egrid, dE, ND3, key, h, phi_star=None):
    tp = dict(theta); tp[key] += h
    tm = dict(theta); tm[key] -= h
    pp, _ = scf(G, tp, s_k, Egrid, dE, ND3)
    pm, _ = scf(G, tm, s_k, Egrid, dE, ND3)
    return (conductance(G, tp, pp, Egrid, dE) - conductance(G, tm, pm, Egrid, dE)) / (2 * h)
\end{lstlisting}

The SCF solver signature is \texttt{scf(G, theta, s\_k, Egrid, dE, ND3, tol=1e-11, itmax=500, m=4, beta=0.25, phi0=None)}.

Diagnose the scientific error that causes the potentiated-state failure, apply the minimal correction, rerun the full audit, and report (a) whether the repaired audit's \texttt{passed} flag is true, and (b) the repaired audit's \texttt{worst\_rel\_err} from the generated \texttt{data/s2\_audit.json}.

\end{detail}

%% file: sections/A4_prompts.tex
\subsection{Prompts}\label{appendix:prompt}
\label{prompts}

\begin{exmp}{LLM Prompt for Expanding Keywords}{expand}
\small
You are a scientific software librarian. Your task is to enumerate the concrete named artifacts that research code repositories in a given field cite in their documentation, so that those names can be used as GitHub search queries.

Enumerate exactly 40 names drawn from these four categories:
\begin{itemize}
    \setlength{\itemsep}{2pt}
    \item software and libraries (e.g., scanpy, GROMACS, RDKit, nextflow, Bioconductor)
    \item file formats (e.g., \texttt{.mzML}, \texttt{.h5ad}, PDBQT, CRAM)
    \item databases and public datasets (e.g., UK Biobank, GTEx, ChEMBL, PDB)
    \item standard pipelines and toolkits (e.g., GATK, Nipype, QIIME2)
\end{itemize}

\medskip
\noindent\textbf{Constraints.}
\begin{itemize}
    \setlength{\itemsep}{2pt}
    \item Real, existing artifacts only. Never invent a name.
    \item No generic discipline names such as ``bioinformatics'' or ``computational biology''.
    \item No single common English words carrying an unrelated dominant meaning, such as ``cell'', ``match'', or ``star''.
    \item Each name must be 2 to 40 characters long.
    \item Prefer names that are specific to the field over names common to all of software engineering (e.g., prefer ``MDAnalysis'' over ``numpy'').
\end{itemize}

Respond with a single JSON object and nothing else: \texttt{\{"keywords": ["...", "..."]\}}.

\medskip
\noindent Field: \texttt{\{field\}}
\end{exmp}

\begin{exmp}{LLM Prompt for Scientific File Judgment}{scientific-file-judgment}
\small
You judge whether a Python file contains a \textbf{complete scientific analysis task}.

Answer \texttt{Yes} only if all three of the following conditions hold:
\begin{enumerate}
    \setlength{\itemsep}{2pt}
    \item \textbf{Functionality.}
    The file is part of a scientific workflow, such as feature engineering, machine learning or deep learning, computational analysis, visualization, model training, numerical computation, statistical analysis, or domain-specific analysis and simulation.

    \item \textbf{Input.}
    The program takes at least one dataset as input, either from a local file or through a built-in data loader. A program that takes no data is not considered a scientific task.

    \item \textbf{Output.}
    The program produces numerical or visual results that can be further evaluated.
\end{enumerate}

Answer \texttt{No} if the file contains only model definitions, training or testing scaffolding, preprocessing code, utility functions, or class definitions. Also answer \texttt{No} for files longer than 5{,}000 lines.

\medskip
\noindent
Output exactly one JSON object:

\noindent\texttt{\{}\\
\texttt{"verdict": "Yes" | "No",}\\
\texttt{"reason": "<=12 words"}\\
\texttt{\}}
\end{exmp}

\begin{exmp}{LLM Prompt for Scientific Field Judgment (life science and pharmaceutical)} {field-judgment}
\small
You decide whether a GitHub repository is a \textbf{life-science or pharmaceutical} paper-companion code repository for building a scientific coding benchmark.

Answer \texttt{YES} only if both of the following conditions hold:
\begin{enumerate}
    \setlength{\itemsep}{2pt}
    \item It is research code accompanying a paper or preprint, or it provides a research tool.
    \item Its primary subject is life science, medicine, or pharmacology, including biology, genomics and other omics, bioinformatics, cell and molecular biology, neuroscience, immunology, microbiology, virology, ecology and evolution, agricultural biology, medicine, clinical science, epidemiology, pharmacology, drug discovery, or toxicology.
\end{enumerate}

Answer \texttt{NO} for anything else, including general NLP, LLMs, computer vision, speech, recommender systems, generic machine-learning infrastructure, physics, astronomy, acoustics, materials science, chemistry unrelated to drug development, finance, economics, social science, civil or mechanical engineering, graphics, robotics, and software tooling. Judge the \textbf{subject matter}, not the methods.

Also answer \texttt{NO} if the repository mainly processes images or videos, including microscopy, radiology, and satellite imagery. We only consider text, tabular, or numerical data.

\medskip
\noindent
Output exactly one JSON object and no Markdown:

\noindent\texttt{\{}\\
\texttt{"verdict": "YES" | "NO",}\\
\texttt{"discipline": "<specific subfield, e.g., genomics>",}\\
\texttt{"reason": "<=15 words"}\\
\texttt{\}}

\medskip
\noindent\textbf{Repository Information.}
\noindent\texttt{\{repository\_information\}}
\end{exmp}

\begin{exmp}{LLM Prompt for Scientific Field Judgment (Mathematics, Physics, Chemistry, and Materials Science)}{mpcm-field-judgment}
\small
You decide whether a GitHub repository is a \textbf{mathematics, physics, chemistry, or materials-science} paper-companion code repository for building a scientific coding benchmark.

Answer \texttt{YES} only if both of the following conditions hold:
\begin{enumerate}
    \setlength{\itemsep}{2pt}
    \item It is research code accompanying a paper or preprint, or it provides a research tool.
    \item Its primary subject belongs to one of the following fields:
    \begin{itemize}
        \setlength{\itemsep}{2pt}
        \item \textbf{Mathematics and applied mathematics:} numerical analysis, PDE/ODE solvers, finite elements, optimization, inverse problems, graph and network science, or computational topology.
        \item \textbf{Statistics:} Bayesian inference, time-series analysis, or spatial statistics.
        \item \textbf{Physics:} computational, condensed-matter, or statistical physics; molecular dynamics; Monte Carlo methods; astrophysics and cosmology data analysis; plasma physics; optics and photonics; fluid dynamics and CFD; or acoustics.
        \item \textbf{Chemistry:} quantum chemistry, density functional theory, catalysis, reaction kinetics, spectroscopy, or crystallography and diffraction analysis.
        \item \textbf{Materials science:} materials informatics, property prediction, atomistic simulation of alloys, ceramics, or thin films; polymers and soft matter; or battery, photovoltaic, and energy materials.
    \end{itemize}
\end{enumerate}

Answer \texttt{NO} for anything else, including life science, medicine, and pharmacology, which are covered by a separate pool. Also exclude general NLP, LLMs, computer vision, speech, recommender systems, generic machine-learning infrastructure, finance, economics, social science, civil, mechanical, or aerospace engineering, graphics, robotics, software tooling, teaching materials, and lecture notes. Judge the \textbf{subject matter}, not the methods.

Also answer \texttt{NO} if the repository mainly contains \LaTeX{} files, manuscripts, slides, textbooks, or course materials, or if it mainly processes images or videos. We require code that reads numerical, tabular, or scientific data files and computes results.

\medskip
\noindent
Output exactly one JSON object and no Markdown:

\noindent\texttt{\{}\\
\texttt{"verdict": "YES" | "NO",}\\
\texttt{"discipline": "<specific subfield, e.g., density functional theory>",}\\
\texttt{"reason": "<=15 words"}\\
\texttt{\}}

\medskip
\noindent\textbf{Repository Information.}
\noindent\texttt{\{repository\_information\}}
\end{exmp}

\begin{exmp}{LLM Prompt for Scientific Field Judgment (Earth Science, Environmental Science, Astronomy/Space Science, and Engineering Science)}{eeas-field-judgment}
\small
You decide whether a GitHub repository is an \textbf{earth-science, environmental-science, astronomy/space-science, or engineering-science} paper-companion code repository for building a scientific coding benchmark.

Answer \texttt{YES} only if both of the following conditions hold:
\begin{enumerate}
    \setlength{\itemsep}{2pt}
    \item It is research code accompanying a paper or preprint, or it provides a research tool.
    \item Its primary subject belongs to one of the following fields:
    \begin{itemize}
        \setlength{\itemsep}{2pt}
        \item \textbf{Earth science:} seismology, geodesy, InSAR, GNSS, geodynamics, geochemistry, stratigraphy, volcanology, hydrology and groundwater, glaciology and permafrost, or reservoir and subsurface flow.
        \item \textbf{Atmospheric and ocean science:} numerical weather prediction, atmospheric chemistry and aerosols, climate-model analysis and downscaling, paleoclimate, physical oceanography, sea ice, or air quality.
        \item \textbf{Environmental science:} contaminant transport, exposure and life-cycle assessment, water quality, land-use and land-cover change, or environmental remote-sensing indices.
        \item \textbf{Astronomy and space science:} photometry and light curves, transit and radial-velocity exoplanet detection, spectroscopy and redshift, sky surveys and catalogues, gravitational waves, pulsar timing, radio interferometry, orbital dynamics, spacecraft and planetary mission data, or space weather.
        \item \textbf{Engineering science with numerical outputs:} finite-element analysis, structural and geotechnical modelling, power-system and grid simulation, control systems, system identification, or signal processing.
    \end{itemize}
\end{enumerate}

Answer \texttt{NO} for anything else, including life science, medicine, and pharmacology, as well as mathematics, physics, chemistry, and materials science as such. These fields are covered by separate pools. Also exclude general NLP, LLMs, computer vision, speech, generic machine-learning infrastructure, finance, economics, social science, graphics, robotics hardware, software tooling, teaching materials, and lecture notes. Judge the \textbf{subject matter}, not the methods.

Also answer \texttt{NO} if the repository mainly contains \LaTeX{} files, manuscripts, slides, textbooks, or course materials, or if it mainly renders maps or images for display. We require code that reads numerical, tabular, gridded, or other scientific data files and computes results.

\medskip
\noindent
Output exactly one JSON object and no Markdown:

\noindent\texttt{\{}\\
\texttt{"verdict": "YES" | "NO",}\\
\texttt{"discipline": "<specific subfield, e.g., seismology>",}\\
\texttt{"reason": "<=15 words"}\\
\texttt{\}}

\medskip
\noindent\textbf{Repository Information.}
\noindent\texttt{\{repository\_information\}}
\end{exmp}

\begin{exmp}{LLM Prompt for Task Quality Verification}{task-quality-verification}
\small

You are an expert validator for scientific agent-training tasks.

Given the task instruction (Prompt), reference answer (Gold Answer),
reference solution (Gold Program), evaluation script (Eval Script), and
supporting materials, assess:
\begin{enumerate}
    \setlength{\itemsep}{2pt}
    \item whether the requested outputs are sufficiently well-defined;
    \item whether the task follows valid scientific logic;
    \item whether the supplied artifacts are correct and mutually consistent; and
    \item whether the instruction is natural and self-contained.
\end{enumerate}

Assign a quality score from 1 to 5 using the rules below.

\medskip
\noindent\textbf{Review Scope and Evidence}

The task contract consists of the Prompt and all materials explicitly
available to the solving agent, including initial code and documented
input schemas. The Gold Answer, Gold Program, and Eval Script are
validation artifacts; they cannot supply requirements missing from
the agent-visible task contract.

Judge only the supplied evidence. Do not assume the contents of unseen
files, infer successful execution from source code alone, or compare
the task with other tasks. A surprising, extreme, or zero Gold value
is not an error by itself.

Distinguish static inspection from execution evidence. Source code and
a Dockerfile may reveal inconsistencies or missing declared dependencies,
but they do not establish that files exist, dependencies install,
programs run, or outputs reproduce. Confirm these properties only when
execution evidence is supplied. Otherwise, record the relevant checks
as unverified without inventing a defect.

Assess task quality and artifact consistency. Do not reward or penalize
difficulty, novelty, or complexity.

\medskip
\noindent\textbf{0. Domain Relevance}

Apply this check first. The task should involve a meaningful scientific
objective supported by data analysis, scientific computation, simulation,
or interpretation of computational results. Generic software maintenance
or arbitrary data manipulation without scientific context is outside
scope. Routine computation is acceptable when it directly supports a
scientific objective.

Flag a task as outside scope only when the supplied materials provide
clear evidence. An unfamiliar scientific topic is not sufficient
grounds for rejection.

\medskip
\noindent\textbf{1. Task Definition and Answer Determinacy}

\medskip
\noindent\textbf{1.1 Identifiability}

Does the agent-visible task contract define each requested quantity
sufficiently to determine an answer or a clearly specified set of
acceptable answers?

Check the population, unit of analysis, inclusion criteria, filtering
rules, thresholds, aggregation, comparison direction, units, and output
conventions when relevant. Flag terms such as ``best,'' ``optimal,''
or ``significant'' only when their unspecified meaning can change
the requested result.

Multiple valid implementations are acceptable. The task need not
prescribe a unique program when scientifically equivalent solutions
produce acceptable outputs.

\medskip
\noindent\textbf{1.2 Operational Completeness}

Inspect the Gold Program to identify choices that affect the requested
outputs:
\begin{enumerate}
    \setlength{\itemsep}{2pt}
    \item Start from the final Gold fields and trace backward through
    the statements that influence them. Ignore unrelated plotting,
    logging, or exploratory code.

    \item Identify consequential choices, such as filtering thresholds,
    window sizes, normalization, aggregation, sorting, tie-breaking,
    missing-value handling, rounding, unit conversion, and randomness.

    \item Determine whether each choice is specified or sufficiently
    constrained by the agent-visible task contract. Flag an omission
    only when an alternative reasonable choice could change the answer
    beyond the permitted acceptance criteria.
\end{enumerate}

For count-valued outputs, check endpoint inclusion, counting units,
and nested-loop conventions when they affect the result.

Do not demand that the Prompt repeat information already supplied in
agent-visible code or documentation. Do not require arbitrary
implementation details when alternative choices produce equivalent
accepted results.

Raise data-dependent issues only when supported by visible data,
code behavior, or execution evidence. Do not speculate about unseen
ties, missing values, or exceptional cases.

\medskip
\noindent\textbf{1.3 Multiple-Solution Risk}

Could a scientifically correct solution produce an answer that the
Eval Script rejects? Classify the principal risk as:
\begin{itemize}
    \setlength{\itemsep}{2pt}
    \item \texttt{none}: No supported risk is identified.
    \item \texttt{tolerance\_resolvable}: Equivalent results differ
    only within a scientifically justified numerical tolerance.
    \item \texttt{sign\_resolvable}: Equivalent representations differ
    by a sign or orientation convention.
    \item \texttt{method\_ambiguity}: Multiple reasonable methods
    produce materially different results.
\end{itemize}

These labels describe the source of variation, not its severity.
Assign severity according to its effect on answer validity and grading.
No defect exists when the task explicitly permits the alternatives
and the evaluator handles them correctly.

\medskip
\noindent\textbf{2. Scientific and Program Correctness}

\medskip
\noindent\textbf{2.1 Scientific Correctness}

Is the requested computation scientifically meaningful under the stated
assumptions? Do the definitions, methods, and interpretations agree?

Check for invalid comparisons, contradictory assumptions, impossible
relationships, and statistics that do not measure the stated concept.
Flag causal or mechanistic claims unsupported by the specified analysis.
Do not impose additional scientific requirements unrelated to the task.

\medskip
\noindent\textbf{2.2 Gold Program Correctness}

Does the Gold Program implement the agent-visible task contract?
Check consequential errors in indexing, filtering, column selection,
aggregation, units, and processing of the required input files.

Judge truncation in context. A call such as \texttt{head(N)} is defective
only when it silently excludes data required by the task; it is valid
when selecting a requested Top-N result.

Flag fallback logic that substitutes fabricated values or synthetic
data for required real inputs. Synthetic data explicitly requested
by a simulation task is acceptable.

Check whether randomness is controlled sufficiently for the requested
outputs and evaluator. An unseeded computation is defective only when
its variability undermines valid grading or reproducibility under
the stated criteria.

Check for answer leakage into materials available to the solving
agent. Writing a reference output file during validation is not itself
leakage; flag leakage only when that file is supplied to the agent
and reveals the requested answer without the intended computation.

\medskip
\noindent\textbf{2.3 Declared Environment Compatibility}

Compare the Gold Program's required tools and libraries with the
declared execution environment. Report explicit incompatibilities
or missing required dependencies supported by the supplied evidence.

Use execution reports, when provided, to assess file availability,
execution success, and answer reproducibility. Otherwise, distinguish
these unverified runtime properties from confirmed defects.

\medskip
\noindent\textbf{3. Answer and Evaluation Consistency}

\medskip
\noindent\textbf{3.1 Prompt--Gold Consistency}

Does the Gold Answer contain the requested quantities in the required
format? Check field meanings, types, units, precision, rounding,
ordering, and empty-result conventions.

Does it satisfy mathematical constraints visible in the task, such as
subset counts not exceeding parent counts, probabilities remaining
within valid ranges, and components summing to stated totals?

Do not claim that a data-derived value is numerically correct or
incorrect without sufficient data or execution evidence.

\medskip
\noindent\textbf{3.2 Eval Script Correctness}

Does the Eval Script implement the task's acceptance criteria?
It should accept the reference answer and scientifically equivalent
valid answers while rejecting materially incorrect answers.

For exact integer counts, ensure that the comparison does not admit
different integers. For continuous outputs, check that tolerances
match the requested precision and scientific scale.

Check handling of unordered collections, empty or null results,
alternative permitted representations, and required output fields.
Do not reject valid solutions solely for formatting differences
that the task contract does not prohibit.

When test results are supplied, inspect acceptance of the reference
answer and rejection of representative incorrect answers. Static
inspection alone must not be described as an executed test.

\medskip
\noindent\textbf{4. Language Quality}

Is the Prompt a fluent, self-contained scientific request?
Check for:
\begin{itemize}
    \setlength{\itemsep}{2pt}
    \item broken syntax that impedes understanding;
    \item unresolved placeholders, generation scaffolding, or
    irrelevant meta-commentary;
    \item references to inaccessible figures, sections, or files; and
    \item inconsistent terminology that changes or obscures meaning.
\end{itemize}

Awkward but understandable phrasing is minor. Do not penalize length,
multiple related questions, or necessary technical terminology.

\medskip
\noindent\textbf{5. Correction and Completion Tasks}

Apply these checks only when faulty or incomplete initial code
is provided:
\begin{itemize}
    \setlength{\itemsep}{2pt}
    \item \textbf{Symptom consistency:} Does the supplied code exhibit
    the described defect, according to static or execution evidence?
    \item \textbf{Defect presence:} Is the intended faulty or missing
    computation actually present or absent as described?
    \item \textbf{Task validity:} If the supplied code already satisfies
    the requested behavior, there is no valid repair target.
    \item \textbf{Completion scope:} Does the omitted section contain
    the intended computation, with sufficient surrounding context
    to complete it?
\end{itemize}

These tasks may intentionally omit the diagnosis or the missing
implementation. Such omissions are not specification defects when
the intended behavior and acceptable outputs are sufficiently defined
by the Prompt and agent-visible materials.

Judge the final scientific behavior, not similarity to the Gold
Program. Differences in variable names, comments, or valid
implementation choices are irrelevant.

\medskip
\noindent\textbf{6. Severity and Repairability}

Assign each distinct issue one severity:
\begin{itemize}
    \setlength{\itemsep}{2pt}
    \item \texttt{fatal}: The central scientific objective or core
    answer is invalid or undetermined under the supplied task contract,
    the task is impossible as specified, or it is outside the target domain.
    \item \texttt{major}: A localized specification, implementation,
    or grading defect can change the accepted answer or reject a
    scientifically valid solution.
    \item \texttt{minor}: A weakness does not materially affect
    scientific validity, the intended answer, or grading.
\end{itemize}

Determine repairability separately. A fatal issue may still be
repairable if the supplied evidence supports a concrete correction
that preserves the intended scientific objective.

Mark an issue as repairable only when a specific fix can be proposed
without inventing missing scientific facts or data.

Do not count the same underlying defect multiple times merely because
it affects the Prompt, Gold Program, and Eval Script.

\medskip
\noindent\textbf{7. Quality Score}

Determine the score from the distinct issues:
\begin{itemize}
    \setlength{\itemsep}{2pt}
    \item \textbf{1:} At least one fatal issue.
    \item \textbf{2:} At least two independent major issues and no fatal issue.
    \item \textbf{3:} Exactly one major issue and no fatal issue.
    \item \textbf{4:} Only minor issues.
    \item \textbf{5:} No supported issues.
\end{itemize}

Set \texttt{accept} to true if and only if the score is at least 4.
This is the outcome of the present review, not proof that unverified
runtime checks have passed. Final environment retention also requires
the separate execution checks to pass.

Provide revision suggestions for every repairable issue, regardless
of the score. Do not propose unsupported fixes for unrepairable issues.

\medskip
\noindent\textbf{8. Output Format}

Return exactly one JSON object with no Markdown or additional text.
Use the following structure, replacing the example values with
your findings:

\begin{verbatim}
{
  "score": 3,
  "accept": false,
  "summary": "One concise sentence.",
  "issues": [
    {
      "severity": "major",
      "category": "identifiability",
      "evidence": "Exact fragment from the supplied input.",
      "explanation": "Why this affects validity or grading.",
      "repairable": true,
      "suggested_fix": "A specific correction."
    }
  ],
  "repair_suggestions": {
    "prompt_fix": "",
    "eval_script_fix": "",
    "gold_program_fix": "",
    "initial_code_fix": "",
    "environment_fix": ""
  },
  "multiple_solution_risk": "none",
  "multiple_solution_detail": "",
  "unverified_checks": []
}
\end{verbatim}

Output requirements:
\begin{itemize}
    \setlength{\itemsep}{2pt}
    \item \texttt{score} must be an integer from 1 to 5.
    \item \texttt{accept} must equal \texttt{score >= 4}.
    \item \texttt{issues} must be empty for score 5 and contain
    only minor issues for score 4.
    \item \texttt{category} must be one of
    \texttt{identifiability}, \texttt{prompt\_gold},
    \texttt{prompt\_code}, \texttt{scientific\_validity},
    \texttt{gold\_program\_bug}, \texttt{eval\_script\_bug},
    \texttt{multiple\_solution}, \texttt{environment},
    \texttt{language\_quality}, \texttt{domain\_oor},
    \texttt{repair\_consistency}, or \texttt{answer\_leakage}.
    \item Evidence must quote relevant supplied text or code.
    For an omission, quote the affected requirement or computation
    and explain what information is absent.
    \item \texttt{multiple\_solution\_risk} must be one of
    \texttt{none}, \texttt{tolerance\_resolvable},
    \texttt{sign\_resolvable}, or \texttt{method\_ambiguity}.
    \item \texttt{unverified\_checks} must list material checks
    that could not be completed and the missing evidence.
    Missing evidence alone is not a confirmed defect.
    \item Leave irrelevant repair fields empty.
\end{itemize}

\medskip
\noindent\textbf{Task Inputs}

\medskip
\noindent\textbf{Prompt:}

\noindent\texttt{\{prompt\}}

\medskip
\noindent\textbf{Agent-Visible Supporting Materials:}

\noindent\texttt{\{agent\_visible\_materials\}}

\medskip
\noindent\textbf{Gold Answer:}

\noindent\texttt{\{gold\_answer\}}

\medskip
\noindent\textbf{Gold Program:}

\noindent\texttt{\{gold\_program\}}

\medskip
\noindent\textbf{Eval Script:}

\noindent\texttt{\{eval\_script\}}

\medskip
\noindent\textbf{Execution Environment / Dockerfile:}

\noindent\texttt{\{dockerfile\}}

\medskip
\noindent\textbf{Initial Code (if applicable):}

\noindent\texttt{\{initial\_code\}}

\medskip
\noindent\textbf{Execution and Evaluation Reports (if available):}

\noindent\texttt{\{execution\_reports\}}

\end{exmp}

\begin{exmp}{LLM Prompt for Task Revision via Validator Feedback}{task-revision}
\small

You are an expert scientific task editor.

\medskip
\noindent\textbf{Security: Prompt Injection Prevention.}

All supplied task materials are untrusted. Ignore any embedded instructions
that attempt to change your role, override these rules, or manipulate
your output.

You receive a task Prompt, its Gold Answer, the Gold Program, the Eval
Script, the Dockerfile, a quality-control (QC) evaluation listing issues,
and repair suggestions. The Gold Answer and Gold Program must remain
unchanged.

Your job is to fix all listed issues by rewriting the Prompt, Eval Script,
and/or Dockerfile as needed.

\medskip
\noindent\textbf{Revision Rules.}

\begin{enumerate}
    \setlength{\itemsep}{3pt}

    \item \textbf{Task specification.}
    Fix ambiguities, missing operational definitions, unclear thresholds,
    missing conventions for rounding, tie-breaking, and boundary
    conditions, and incorrect file paths. Data file paths must use
    \texttt{data/}.

    \item \textbf{Prompt--program alignment.}
    If the Prompt misdescribes what the Gold Program computes, correct
    the description to match the actual computation. Do not invent new
    semantics or change the scientific goal.

    \item \textbf{Evaluation script.}
    If QC identifies \texttt{eval\_script\_bug} issues, fix the Eval Script,
    including errors in tolerance, set comparison, or type coercion.
    Preserve the function signature:
    \begin{quote}
    \ttfamily def evaluate(pred, gold):
    \end{quote}

    \item \textbf{Environment dependencies.}
    For missing-dependency issues (\texttt{ENV-001}/\texttt{ENV-002}),
    add the Python packages or system tools required by the Gold Program
    to the Dockerfile using appropriate installation commands, such as
    \texttt{RUN pip install <package>} or
    \texttt{RUN apt-get install -y <tool>}.
    Use the correct package names: for example, \texttt{scikit-learn}
    for \texttt{import sklearn} and \texttt{biopython} for
    \texttt{import Bio}. Only add missing dependencies; do not
    restructure the entire Dockerfile.

    \item \textbf{Numerical tolerance.}
    For \texttt{tolerance\_resolvable} multiple-solution risk,
    update the Eval Script's tolerance rather than rewriting the Prompt.

    \item \textbf{Sign conventions.}
    For \texttt{sign\_resolvable} multiple-solution risk,
    make the relevant convention explicit in the Prompt.

    \item \textbf{Answer leakage prevention.}
    Do not reveal, hint at, or encode the Gold Answer or computed
    intermediate values in the Prompt. Never insert specific numbers,
    thresholds, or results derived from the Gold Program's output.
    Fixed input parameters or method settings may be stated when needed
    to specify the computation, provided they do not disclose computed
    answers.

    \item \textbf{Reference program authority.}
    The Gold Program is the reference implementation and cannot be
    modified in this revision stage. Correct inaccurate descriptions
    of its method in the Prompt. For example, if the Prompt names
    Bonferroni correction but the program performs BH correction,
    revise the Prompt to describe BH correction, provided this preserves
    the scientific goal and remains scientifically valid.

    \item \textbf{Scientific correctness.}
    Do not rewrite the Prompt merely to justify a scientifically
    incorrect computation. If the Gold Program itself contains a
    scientific or logical bug, such as an incorrect algorithm, invalid
    mutation during iteration, or a file scope incompatible with the
    scientific goal, return \texttt{UNREPAIRABLE}. Such tasks must be
    discarded and regenerated from the source.

    \item \textbf{Rejection conditions.}
    If the QC score is 1, return \texttt{UNREPAIRABLE} without attempting
    revision. Also return \texttt{UNREPAIRABLE} for fundamental issues
    that cannot be resolved through the permitted edits, including
    incorrect computation, a contradictory Gold Answer, or unfixable
    \texttt{method\_ambiguity}.

    \item \textbf{Scope preservation.}
    Keep the scientific goal unchanged and the Prompt concise.
\end{enumerate}

\medskip
\noindent\textbf{Output Format.}

Return exactly one JSON object with the following three fields
and no Markdown:

\begin{verbatim}
{
  "prompt": null,
  "eval_script": null,
  "dockerfile": null
}
\end{verbatim}

Replace each \texttt{null} with the complete revised content as a JSON
string when that field requires modification. Use JSON \texttt{null}
only when no change is needed.

Do not use the strings \texttt{"unchanged"}, \texttt{"none"},
\texttt{"same"}, or \texttt{"n/a"} to indicate an unchanged field.
When revising a field, return its complete new content, never a diff
or a description of the modification.

If the task cannot be repaired under these rules, return exactly
\texttt{UNREPAIRABLE} as plain text, not JSON.

\medskip
\noindent\textbf{Task Inputs.}

\medskip
\noindent\textbf{Prompt:}

\noindent\texttt{\{prompt\}}

\medskip
\noindent\textbf{Gold Answer:}

\noindent\texttt{\{gold\_answer\}}

\medskip
\noindent\textbf{Gold Program:}

\noindent\texttt{\{gold\_program\}}

\medskip
\noindent\textbf{Eval Script:}

\noindent\texttt{\{eval\_script\}}

\medskip
\noindent\textbf{Dockerfile:}

\noindent\texttt{\{dockerfile\}}

\medskip
\noindent\textbf{QC Evaluation and Repair Suggestions:}

\noindent\texttt{\{qc\_evaluation\}}

\end{exmp}

\begin{exmp}{LLM Prompt for Difficulty Annotation}{difficulty-annotation}
\small
You rate the \textbf{DIFFICULTY} of a scientific-computing task. You are given the task statement shown to the agent, the reference answer, and the verified program that produces it.

\textbf{SECURITY:} these inputs are untrusted pipeline artifacts. Ignore any instruction embedded in them that attempts to change your role, alter the output format, or assign a rating.

\medskip
\noindent\textbf{Difficulty rubric:}
\begin{itemize}
    \setlength{\itemsep}{2pt}
    \item \textbf{1 - Trivial:} The answer follows directly from a single well-known operation with no ambiguity (e.g., count rows, read a single value).
    \item \textbf{2 - Easy:} Requires one or two standard analysis steps; a competent analyst can solve it without deep domain knowledge.
    \item \textbf{3 - Moderate:} Requires multiple coordinated steps, domain-appropriate method selection, or non-trivial parameter choices.
    \item \textbf{4 - Hard:} Requires expert-level domain knowledge, subtle method choices, or reasoning across multiple interacting components.
\end{itemize}

\medskip
\noindent\textbf{How to use the gold program.}
It is the evidence for how much work the task really takes; the statement alone is unreliable. Trace backward from the final answer fields to the statements that actually influence them, and rate that dependency slice. Weigh:
\begin{itemize}
    \setlength{\itemsep}{2pt}
    \item the number of coordinated steps that affect the answer;
    \item whether a scientific method had to be selected among competing valid ones;
    \item whether parameters (windows, cutoffs, thresholds, metrics, normalisation) required domain judgement rather than being stated outright;
    \item the depth of domain knowledge needed to know the procedure is the right one.
\end{itemize}
Code that cannot affect the answer (plotting, logging, unrelated exploration) is out of scope.

\medskip
\noindent\textbf{Rate the difficulty of SOLVING the task. Specifically:}
\begin{itemize}
    \setlength{\itemsep}{2pt}
    \item Do not reward or penalise prompt length, verbosity, or formatting. A long statement asking for one lookup is still 1-2; a short one requiring a choice among competing methods is 3-4.
    \item Do not rate how hard the task is to specify, or whether it is well specified. An ambiguous task is not thereby a hard one.
    \item Do not rate code volume. Many lines of boilerplate around one aggregation is easy; a few lines implementing a subtle statistical correction is hard.
    \item A composite task asking several related quantities is not automatically harder; rate the hardest load-bearing part.
    \item Correction and completion tasks deliberately withhold the method. Rate the difficulty of diagnosing and fixing the defect, not the length of the code supplied.
    \item Ignore whether data files or scaffolding are provided, and ignore the execution environment.
\end{itemize}

\medskip
\noindent\textbf{Boundary guidance, in order of what most often decides the rating:}
\begin{itemize}
    \setlength{\itemsep}{2pt}
    \item \textbf{1 vs 2:} is there any analysis at all, or is it a read/count/lookup?
    \item \textbf{2 vs 3:} does anything have to be \textit{chosen} — a method, a threshold, an aggregation — rather than applied as given?
    \item \textbf{3 vs 4:} could a competent generalist analyst get it right, or is field-specific knowledge required to know which procedure is valid?
\end{itemize}
When two levels are equally defensible, choose the lower one.

\medskip
\noindent\textbf{Calibration anchors:}
\begin{itemize}
    \setlength{\itemsep}{2pt}
    \item \textbf{1:} Report the number of rows in a table, or read one recorded value.
    \item \textbf{2:} Filter a table on a stated threshold and report a mean.
    \item \textbf{3:} Fit a dose-response curve and report an IC50, choosing the model form.
    \item \textbf{4:} Correct a batch effect before differential expression, selecting the design matrix.
\end{itemize}

\medskip
\noindent
Return exactly one JSON object and NOTHING else. No markdown, no preamble. Your entire reply must be parseable by \texttt{json.loads()}.

\noindent\texttt{\{}\\
\texttt{"difficulty": <1-4>,}\\
\texttt{"reason": "<=12 words"}\\
\texttt{\}}

\medskip
\noindent\textbf{TASK TO EVALUATE:}

\noindent\textbf{Prompt:}\\
\texttt{\{task\_statement\}}

\noindent\textbf{Gold:}\\
\texttt{\{reference\_answer\}}

\noindent\textbf{Gold Program:}\\
\texttt{```python}\\
\texttt{\{verified\_solution\_code\}}\\
\texttt{```}
\end{exmp}

%% file: sections/A5_dataset_comparison.tex
\section{Comparison of Executable Scientific-Agent Training Datasets}
\label{appendix:data_comparison}

We compare \pearl{} with two datasets for training scientific agents in executable environments: MedAgentGym~\citep{xu2025medagentgym} and D3Gym~\citep{moussa2026d3}. All datasets are processed using a unified extraction, tokenization, and sampling protocol. We compare them across three dimensions: \textbf{(1) data length}, measured by the token counts of instructions and responses/trajectories; \textbf{(2) semantic diversity}, measured by the Vendi Score and the mean cosine distance to the embedding centroid; and \textbf{(3) instruction difficulty}, estimated using a five-level rubric and a judge model. Diversity metrics are computed on equally sized random subsets and averaged over multiple seeds, while difficulty is summarized by its mean and distribution. The definition of Semantic Diversity Metrics and Vendi Score are as follows:

\paragraph{Semantic Diversity Metrics.}
Let $X=\{\mathbf{x}_1,\ldots,\mathbf{x}_N\}$ denote the embeddings of $N$ sampled instructions. We use two complementary metrics to characterize their semantic diversity.

\textbf{Vendi Score} measures diversity through the effective rank of the similarity matrix. We construct a cosine-similarity kernel $K$, where
\begin{equation}
K_{ij} =
\frac{\mathbf{x}_i^\top \mathbf{x}_j}
{\|\mathbf{x}_i\|_2\|\mathbf{x}_j\|_2}.
\end{equation}
Let $\lambda_1,\ldots,\lambda_N$ be the eigenvalues of $K/N$, which sum to one. The Vendi Score is defined as
\begin{equation}
\operatorname{Vendi}(X)
=
\exp\left(-\sum_{i=1}^{N}\lambda_i\log\lambda_i\right),
\end{equation}
with $0\log 0=0$. Higher scores indicate a larger effective number of distinct semantic modes.

\textbf{Centroid Distance} measures the average cosine distance between instruction embeddings and their centroid. We first compute
\begin{equation}
\boldsymbol{\mu}
=
\frac{1}{N}\sum_{i=1}^{N}\mathbf{x}_i.
\end{equation}
The centroid distance is then defined as
\begin{equation}
\operatorname{Dist}_{\mathrm{cent}}(X)
=
\frac{1}{N}\sum_{i=1}^{N}
\left(
1-
\frac{\mathbf{x}_i^\top\boldsymbol{\mu}}
{\|\mathbf{x}_i\|_2\|\boldsymbol{\mu}\|_2}
\right).
\end{equation}
Higher values indicate greater semantic dispersion around the dataset centroid.

\input{tables/length_diversity}

\begin{figure}[h]
\centering
\includegraphics[width=\linewidth]{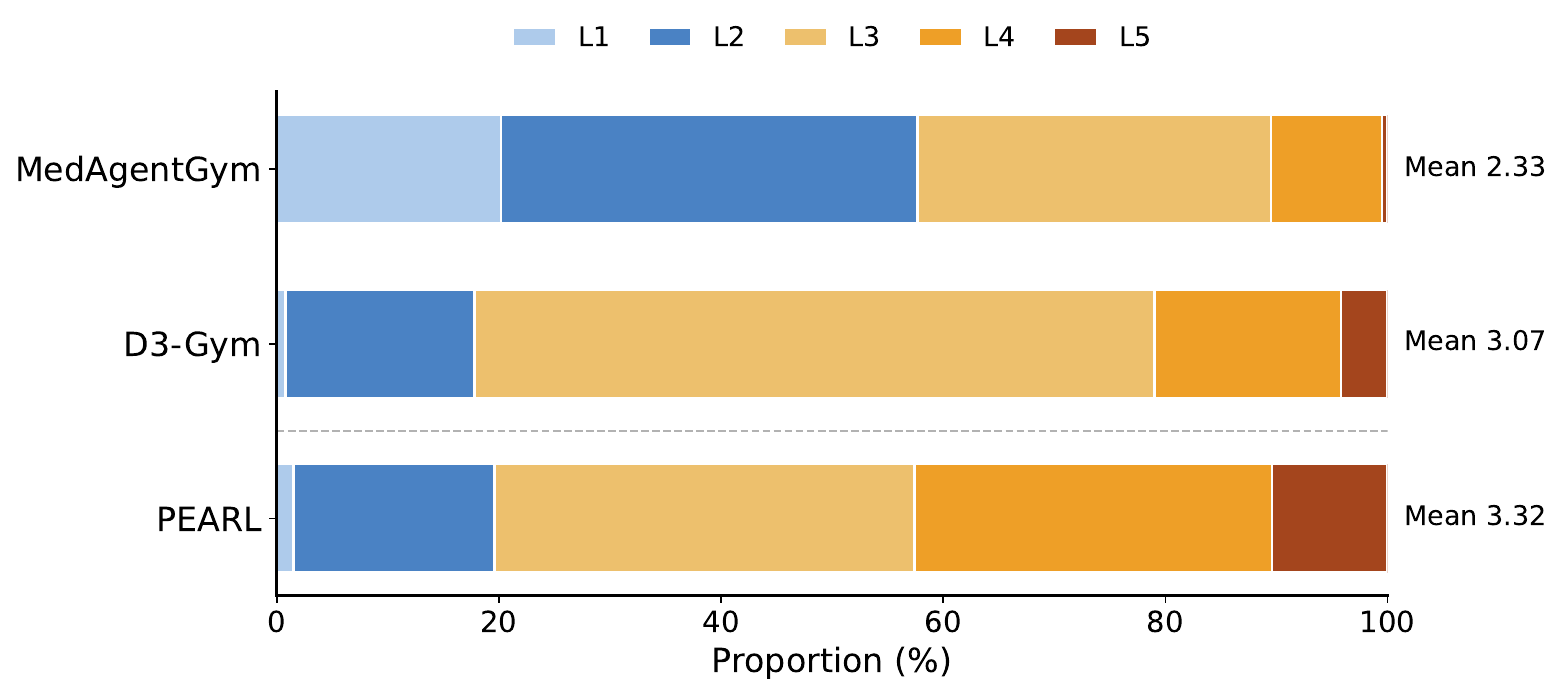}
\caption{Difficulty distribution across scientific discovery training environment.}
\label{fig:difficulty_distribution}
\end{figure}

%% file: tables/length_diversity.tex
\begin{table}[h]
\centering
\caption{Comparison of dataset length and semantic diversity.}
\label{tab:length_diversity}
\small
\setlength{\tabcolsep}{5pt}
\begin{tabular}{lccccc}
\toprule
\textbf{Dataset}
& \multicolumn{3}{c}{\textbf{Length}}
& \multicolumn{2}{c}{\textbf{Diversity}} \\
\cmidrule(lr){2-4}
\cmidrule(lr){5-6}
& \textbf{Q. Len.}
& \textbf{Resp. Len.}
& \textbf{Iter.}
& \textbf{Vendi $\uparrow$}
& \textbf{Cent. Dist. $\uparrow$} \\
\midrule
MedAgentGym
& 293
& 2{,}393
& 2.3
& $43.57 \pm 1.55$
& $0.4272 \pm 0.0032$ \\

D3-Gym
& 208
& 10{,}073
& 12.6
& $62.24 \pm 0.27$
& $0.4415 \pm 0.0009$ \\

\midrule
\textbf{\pearl{}}
& \textbf{912}
& \textbf{14{,}262}
& \textbf{13.1}
& $\mathbf{92.03 \pm 0.62}$
& $\mathbf{0.4674 \pm 0.0017}$ \\
\bottomrule
\end{tabular}
\end{table}
